%% file: alignment_aistats.tex
\documentclass[twoside]{article}

\usepackage[accepted]{aistats2019}
\usepackage[log]{snapshot}

\usepackage[utf8]{inputenc} % allow utf-8 input
\usepackage[T1]{fontenc}    % use 8-bit T1 fonts
\usepackage{hyperref}       % hyperlinks
\usepackage{url}            % simple URL typesetting
\usepackage{booktabs}       % professional-quality tables
\usepackage{amsfonts}       % blackboard math symbols
\usepackage{nicefrac}       % compact symbols for 1/2, etc.
\usepackage{microtype}      % microtypography

\usepackage[T1]{fontenc}
\usepackage{lmodern}

\usepackage{times}

\usepackage[dvipsnames]{xcolor}
\usepackage{color}

\definecolor{redcolor}{RGB}{219, 0, 0}
\definecolor{bluecolor}{RGB}{0, 0, 219}

\usepackage{placeins}

\usepackage{tikz}
\usetikzlibrary{positioning}
\usepackage{algorithm}
\usepackage{algorithmic}

\usepackage[subtle,bibbreaks=normal,mathspacing=normal]{savetrees}

\usepackage{subcaption} % To use subfigures with subcaptions
\usepackage[font=small]{caption}
\usepackage[font=small]{subcaption}
\usepackage{amsmath}
\usepackage{amsfonts}

\usepackage[colorinlistoftodos,prependcaption,textsize=tiny]{todonotes}

\usepackage{resizegather}
\usetikzlibrary{bayesnet}
\usepackage{rotating}

\usepackage{easytable}
\usepackage[toc,page]{appendix}
\usepackage{float}
\usepackage{resizegather}
\usepackage{array}
\usepackage{balance}
\usepackage{flushend}

\graphicspath{{images/}}

\usepackage{xspace}
\makeatletter
\DeclareRobustCommand\onedot{\futurelet\@let@token\@onedot}
\def\@onedot{\ifx\@let@token.\else.\null\fi\xspace}
\def\eg{\emph{e.g}\onedot} 
\def\ie{\emph{i.e}\onedot} 
 
\def\etc{\emph{etc}\onedot} 
 
\def\etal{\emph{et al}\onedot}
\makeatother

\usepackage{enumitem}

\usepackage{titlesec}
\titlespacing*{\paragraph}{0pt}{0.3ex plus .2ex minus .2ex}{1em}
\newcommand{\Y}{\mathbf{Y}}

\newcommand{\p}[2][]{p_{#1}\!\left(#2\right)}

\newcommand{\NormalDistrib}[1]{\mathcal{N}\!\left(#1\right)}

\newcommand{\Yjobs}{{Y}_{j}}
\newcommand{\betaj}{\beta_{j}}
\newcommand{\thetaj}{\theta_{j}}

\newcommand*\rfrac[2]{{}^{#1}\!/_{#2}}

\begin{document}

% If your paper is accepted and the title of your paper is very long,
% the style will print as headings an error message. Use the following
% command to supply a shorter title of your paper so that it can be
% used as headings.
%
%\runningtitle{I use this title instead because the last one was very long}

% If your paper is accepted and the number of authors is large, the
% style will print as headings an error message. Use the following
% command to supply a shorter version of the authors names so that
% they can be used as headings (for example, use only the surnames)
%
%\runningauthor{Surname 1, Surname 2, Surname 3, ...., Surname n}

\twocolumn[

\aistatstitle{Gaussian Process Latent Variable Alignment Learning}

\aistatsauthor{ Ieva Kazlauskaite \And Carl Henrik Ek \And Neill D. F. Campbell}

\aistatsaddress{ University of Bath, UK \\ Electronic Arts \\ \small{\texttt{i.kazlauskaite@bath.ac.uk}} \And  University of Bristol,  UK \And  University of Bath, UK \\ Royal Society \\\small{\texttt{n.campbell@bath.ac.uk}}} 
]

\begin{abstract}
  We present a model that can automatically learn alignments between high-dimensional data in an unsupervised manner. Our proposed method casts alignment learning in a framework where both alignment and data are modelled simultaneously. Further, we automatically infer groupings of different types of sequences within the same dataset. We derive a probabilistic model built on non-parametric priors that allows for flexible warps while at the same time providing means to specify interpretable constraints. We demonstrate the efficacy of our approach with superior quantitative performance to the state-of-the-art approaches and provide examples to illustrate the versatility of our model in automatic inference of sequence groupings, absent from previous approaches, as well as easy specification of high level priors for different modalities of data.
\end{abstract}

\input{includes/intro.tex}
\input{includes/background.tex}
\input{includes/methodology.tex}
\input{includes/experiments.tex}
\input{includes/conclusions.tex}

%\FloatBarrier

\subsubsection*{Acknowledgments}
This work has been supported by EPSRC CDE (EP/L016540/1) and CAMERA (EP/M023281/1) grants as well as the Royal Society. IK would like to thank the Frostbite Physics team at EA.

%\balance
% \newpage
\bibliography{references}
\bibliographystyle{plain}

\newpage
\section*{Supplementary material}

\input{includes/appendix_old.tex}

\end{document}

%% file: includes/intro.tex
% !TEX root = ../gplvmalignICML2018.tex
\section{Introduction}
\label{intro}
Learning from sequential data is challenging as data might be sampled at different and uneven rates, sequences might be collected out of phase, \etc. 
Consider the following scenarios: humans performing a task may take more or less time to complete parts of it, climate patterns are often cyclic though particular events take place at slightly different times in the year, the mental ability of children varies depending on their age, neuronal spike waveforms contain temporal jitter, replicated scientific experiments often vary in timing. 
However, most sample statistics, \eg~mean and variance, are designed to capture variation in amplitude rather than phase/timing. This leads to increased sample variance, blurred fundamental data structures and an inflated number of principal components needed to describe the data. Therefore, the data needs to be aligned in order for dependencies such as these to be recovered. This is a non-trivial task that is often performed as a pre-processing stage to modelling. 

Traditionally, the notion of sequence similarity comes from a measure of pairwise similarity integrated across the sequences. This local measure often leads to highly non-convex optimisations problems making alignments challenging to learn. In this paper we take a different approach where we encapsulate alignment and modelling within a single framework. By simultaneously modelling the sequences and the alignment we can capture global structure thereby circumventing the difficulties associated with an objective function based on pairwise similarity.

Further difficulties arise when the the dataset contains observations from several distinct functions. Consider, for example, a set of motion capture experiments that include tasks such as running, jumping and sitting down. Data for each of these three types of sequences can be aligned to themselves but a global alignment between them may not exist. In traditional approaches, the observed data must be grouped into the distinct sequence types before alignment. To overcome this limitation, our approach also produces a generative model over the sequences themselves. This means we can simultaneously infer both alignments and their grouping.

Methods for learning alignments can broadly be classified into two categories. The first learns a function to warp the input dimension while the second directly learns the transformed sequences. There are several benefits to learning a warping function as it allows us to resample the data and, by constraining the class of functions, we can also incorporate global constraints on the alignment. However, specifying a parametric function is challenging and often results in difficult optimisation tasks.
Directly learning transformed sequences avoids having to specify a parametrisation. However, this comes at the cost of removing all but the most rudimentary global constraints on the warping function since the optimal alignment is completely specified by the pairwise similarity. 
In contrast, we propose a novel approach that learns the warping function using a probabilistic model. Underpinning our methodology is the use of Gaussian process priors that allows us to approach this learning in a Bayesian framework achieving principled regularisation without reducing the solution space.

Our proposed model overcomes a number of problems with the existing literature and confers three main contributions:
\begin{enumerate}[topsep=-3pt,itemsep=-3pt,partopsep=0pt]
\item We model the observed data directly with a generative process, rather than interpolating between observations, that allows us to reject noise in a principled manner.
\item The generative model of the aligned data allows a fully unsupervised approach that performs simultaneous clustering and alignment.
\item We use continuous, non-parametric processes to model explicitly the warping functions throughout; this allows the specification of sensible priors rather than unintuitive or heuristic choices of parametrisations.%
\end{enumerate}

%% file: includes/background.tex
% !TEX root = ../gplvmalignICML2018.tex
\section{Background}
\label{sec:background}

%\subsection{Related work}
\label{sec:related_work}

\paragraph{Pairwise similarity}There has been a significant amount of work in learning alignments from data. Most approaches are based on the assumption of the existence of a pairwise similarity measure between the instances of each sequence. 
%The idea is then to find an alignment of the two sequences such that the sum of these distances is minimised. 
The classical approach to minimise the distance between two sequences is called Dynamic Time Warping (DTW), and is based on a computing an affinity matrix 
% of size $N \times M$ where $N$ and $M$ are the lengths 
of the two sequences to be aligned~\cite{Berndt:1994}. The solution corresponds to the path through this matrix that leads to the minimal combined pairwise cost. The optimal solution is found by backtracking through the affinity matrix and can be estimated using Dynamic Programming~\cite{Muller:2007}. DTW finds the optimal alignment based on a pairwise distance between each element in two sequences. Such formulation imposes a number of limitations. DTW returns an alignment but not a parametrised warping, and it is not trivial to encode a preference towards different warps as this would be a global characteristic while DTW is a local algorithm. 
\paragraph{Multiple sequences}In its original form DTW aligns two sequences only but several extensions allow it to process multiple sequences at once, most notably Procrustes dynamic time warping (PDTW), Procrustes derivative dynamic time warping (PDDTW), and Iterative Motion Warping (IMW)~\cite{Keogh:2001, Drydmard:2016, Hsu:2005}. All of these methods are applied directly in the observation space which is a limitation when the data contains a significant amount of noise. The main algorithms that address this limitation are Canonical Time Warping (CTW) and Generalized Time Warping (GTW)~\cite{Zhou:2009, Zhou:2012}. Both of these approaches perform feature extraction and find a subspace that maximises the linear correlation of data samples. Similarly to our approach, GTW is parametrised using monotonic warping functions. However, in all these methods the spatial alignment and time warping are coupled. Another extension, called Generalized Canonical Time Warping (GCTW) combines CCA with DTW to simultaneously align multiple sequences of multi-modal data~\cite{Zhou:2016}. GCTW relies on additional heuristic energy terms and on coarse-to-fine optimisation to get the energy method to converge to a good local minimum. 

\paragraph{Feature extraction}More recently, deep neural networks were employed to perform temporal alignments~\cite{Trigeorgis:2016}, ~\cite{Trigeorgis:2017}. The proposed method, called Deep Canonical Time Warping (DCTW), performs non-linear feature extraction and it performs competitively on larger audio-visual datasets.  A different method proposed by ~\cite{Listgarten:2004} uses continuous hidden Markov models, where the latent trace is an underlying representation of the set of observable sequences. \cite{Haxby:2011} introduced hyperalignment that finds isometric transformations of trajectories in voxel space that result in an accurate match of the time-series data. An extension to this model was proposed by~\cite{Lorbert:2012} who address the issues of scalability and feature extension through the use of the kernel trick. The authors note that classification accuracy relies on intelligent feature selection. 

\paragraph{Manifold alignment}Similar to our approach,~\cite{Cui:2014} propose an unsupervised manifold alignment method. It is based on finding alignment by enforcing several constraints such as geometry structure and feature matching, geometry preservation and integer constraints. The approach shows promising results but is very computationally expensive. Another non-linear feature extraction method~\cite{Vu:2012} named Manifold Time Warping relies on constructing a k-nearest neighbour graph and then performing DTW to align a pair of sequences. 

\paragraph{Implicit transformation}Another approach to alignment is to use an implicit transformation of the sequences. In \cite{Cuturi:2007,Cuturi:2011} the authors propose a kernel function that is capable of mapping sequences of different length to an implicit feature space. Another similar approach is \cite{Baisero:2015tm} which describes a range of different kernels on sequences, this method is flexible and allows for learning implicit feature space mappings for sequences of not only different lengths but also different dimensionality. These methods work well experimentally but as the alignment is implicit we cannot re-align sequences or construct novel ones.

\paragraph{Shape analysis}A different line of work, often referred to as elastic registration or shape analysis is considered in the functional data analysis literature. In \cite{Garreau:2014} the authors propose an extension to DTW by replacing the Euclidean distance with a Mahalanobis distance. By having a parametrisable distance function the authors are able to learn the metric function from a set of paired observations. \cite{Kurtek:2011} study the group-theoretic approach to warps by using the group of warping functions to describe the equivalence relation between signals. In particular, the authors use the Fisher-Rao Riemannian metric and the resulting geodesic distance to align signals with random warps, scalings and translations. Square root velocity function (SRVF) facilitates the use of Fisher-Rao distance between functions by estimating the $\mathbb{L}^2$ norm between their SRVFs~\cite{Srivastava:2011, Kurtek:2012}. \cite{Tucker:2013} proposed a generative model that combines elastic shape analysis of curves and functional principal component analysis (fPCA). Another recent extension called Elastic functional coding relies on trajectory embeddings on Riemannian manifolds and results in manifold functional variant of PCA~\cite{Anirudh:2015}. 

\paragraph{Gaussian processes}\label{sec:gaussian_processes}%
Our model makes use of Gaussian processes as priors over warpings, sequences and their groupings. A Gaussian process (GP)~\cite{Rasmussen:2005} is a random process specified by a mean $m(x)$ and a covariance function $k_{\theta}(x,x')$. The covariance function is parametrised by a set of hyper-parameters $\theta$ while the mean is often considered as constant zero. The index set of the two functions is infinite which allows GPs to be interpreted as non-parametric priors over the space of functions. Even though the process is infinite, an instantiation of the process is finite and reduces to a Gaussian distribution. In a regression setting we observe a set of noisy samples $\mathcal{D} = \{x_{n}, y_n\}_{n=1}^N$ of a latent function $f(\cdot)$ such that $y_{n} = f(x_{n}) + \varepsilon_{n}$. By placing a GP prior over the latent function $f\sim \mathcal{GP}(m(x),k(x,x'))$ the instantiations of the function at the training data $\{f_n=f(x_n)\}_{n=1}^N$ are Gaussian as $F\sim\mathcal{N}(m(X),k(X,X))$ where $F$ and $X$ are concatenations of the function instantiations and the observed input locations respectively. By choosing a Gaussian noise model the functions can be marginalised out in closed form due to the self-conjugate property of the Gaussian distribution. The Gaussian process latent variable model (GP-LVM)~\cite{Lawrence:2005vk} is a model that uses GP priors to learn latent variables. The model assumes that each dimension of the observed data $Y$ have been generated from a latent variable $X$ through some latent function $f$. By placing a GP prior over $f$ and marginalising out this mapping, the latent representation of the data can be recovered. The model is very flexible and has been implemented across a wide range of different applications, for example~\cite{Campbell:2014, Grochow:2004, Urtasun:2005}.

\paragraph{Warped GPs} In~\cite{Snelson:2004} and~\cite{Lazaro:2012} the authors construct a GP with a warped input space to account for differences in observations (e.g.~inputs may vary over many orders of magnitude), and show that a warped GP finds the standard preprocessing transforms, such as the logarithm, automatically. In comparison, our approach leads to a warped output space of the GP-LVM, and uses the additional knowledge of possible misalignments in the high-dimensional space to regularise the problem of building a low-dimensional latent space. Concurrent with our work,~\cite{Duncker:2018} use GPs for modelling sequences of neural population spike-trains and the corresponding temporal warps. The proposed approach is an extension to GP factor analysis~\cite{Yu:2009} and uses a linear combination of shared and private latent processes to encourage alignment of sequences for different trails. Unlike our work,~\cite{Duncker:2018} do not recover a clustering of the sequences and thus require the groups of sequences for alignment to be known a-priori.
% -----------------------------------------------------------------------------------------

%% file: includes/methodology.tex
% !TEX root = ../alignment_aistats.tex
\section{Methodology}
\label{sec:methodology}

%\subsection{Our model}
\label{sec:our_model}

\input{includes/overview_figure.tex}

Alignment learning is the task of recovering a set of monotonic warping functions that have been used to create samples of a latent sequence. 
Fig.~\ref{fig:overview} provides an overview of our approach. We are provided with a number of noisy time warped observations of a set of unobserved latent sequences and our task is to infer both this set of sequences and the time warps that give rise to the observations.

Let us assume that we have $J$ noisy sequence observations $\{\Yjobs\}$ (Fig.~\ref{fig:overview_obs}) where each observed sequence comprises $N$ time samples, $\Y = (y_{(j,n)}) \in \mathbb{R}^{J \times N}$. We consider each sequence to be generated as a sample from a latent function $f_j(x)$ (Fig.~\ref{fig:overview_fit}) under a monotonic warping $g_j(x)$ as $y_{(j,n)} = f_j( \, g_j(x_n) ) + \varepsilon_{jn}$ where the samples have been corrupted by additive Gaussian noise $\varepsilon_{jn} \sim \mathcal{N}(0, \beta_j^{-1})$. Due to the close association of sequences and temporal data, we use the word time to refer to the input domain of the sequence, however our method is general and applicable to any ordered index set.

The aligned sequences, which are unobserved, are given by the corresponding functions without the time warp $s_{(j,n)} = f_j( x_n ) + \varepsilon_{jn},  \, \mathbf{S} = (s_{(j,n)}) \in \mathbb{R}^{J \times N}$ (as illustrated in Fig.~\ref{fig:overview_true} and Fig.~\ref{fig:overview_aligned}).
This means that we can encode our warping function as the transformation from a \emph{known} sampling of an \emph{unknown} aligned sequence to the \emph{unknown} sampling of the \emph{known} observations for each sequence. However, as described in the introduction and illustrated in Fig.~\ref{fig:overview_true}, we wish to design a model that is not restricted to the case where all the observations arise from a \emph{single} latent function (for example, in Fig.~\ref{fig:overview_true} there are two unknown true sequences). 

To account for the possible existence of multiple latent functions, we consider a generative model for the aligned sequences themselves. By specifying that the generative process be as simple as possible, we encourage the clustering of these sequences, which allows to automatically find the smallest number of latent functions explaining the data. We encode this as the aligned sequences being generated via a smooth mapping $h(\cdot)$ from a low dimensional space $Z \in \mathbb{R}^{Q}$ as
\begin{equation}
S_j = h(Z_j) + \hat{\epsilon} \quad \text{s.t.} \quad  S_j = f_j(x) + \epsilon_j \: \forall \; j \; \label{eqn:such_that}
\end{equation} where $\hat{\epsilon} \sim \NormalDistrib{0, \gamma^{-1} I}$ and $\epsilon_j \sim \NormalDistrib{0, \beta_j^{-1} I}$.
This low dimensional manifold is visualised, for our toy example, in Fig.~\ref{fig:overview_manifold} where $Z$ is a 2D space and the locations of the aligned sequences $S_j$ are shown as coloured points matching the corresponding aligned sequences in Fig.~\ref{fig:overview_all_aligned}. We see that the two different sequences are clustered appropriately by their location in the manifold and the sequences are correctly aligned. We use a probabilistic model of the aligned sequences, which allows us to quantify uncertainty of the low-dimensional manifold representations (the heatmap in Fig.~\ref{fig:overview_manifold}).

\subsection{Probabilistic Model}

In this section we specify the two components of our model that correspond to the constraint introduced in Eq.~\eqref{eqn:such_that}. 

The first part corresponds to fitting the data that explains the observed sequences and specifies the latent functions, while the second part enforces a simple, low-dimensional structure of the aligned sequences. Given noisy observed data, we do not impose this constraint exactly, but rather define both model components as probabilistic models and interpret this constraint as one of the aligned sequences having high likelihood under both model components simultaneously. If the aligned sequences $\mathbf{S}$ were known, this interpretation would correspond exactly to fitting a model to the observed data by maximising the data likelihood. 
Since $\mathbf{S}$ are unobserved, we refer to them as \emph{pseudo-observations}; similarly to~\cite{Titsias:2009}, we augment the probability space with a set of pseudo-observations which are constrained by the two components of our model. We then fit the model by maximising the joint marginal likelihood of observations $\mathbf{Y}$ and pseudo-observations $\mathbf{S}$, while optimising not only w.r.t.~the model parameters, but also w.r.t.~the pseudo-observations $\mathbf{S}$. 

\paragraph{Model over time} We have $Y_j \in \mathbb{R}^{N}$ as the observed sequences, and let $X \in \mathbb{R}^{N}$ denote an observed uniform sampling of time. We introduce a random variable $G_j \in \mathbb{R}^{N}$ to encode the time warp function sampled at $X$ such that $G_j \sim g_j(X)$. The random variables for the functions $f_j(\cdot)$ are more involved since the functions are evaluated at different locations. Let $F_j^{\mathrm{G}} \sim f_j(G_j)$ denote the output of the function sampled at the time warped locations $G_j$ and let $F_j^{\mathrm{X}} \sim f_j(X)$ denote the function evaluated at the uniform sample locations $X$. The observations $Y_j$ are the noise-corrupted versions of $F^G$, and similarly, we call the noise-corrupted version of $F^X$ pseudo-observations, since they are not observed and should be inferred.

We now define the priors over the generating and warping functions $f_j(\cdot)$ and $g_j(\cdot)$. Specifying a parametric mapping is challenging and it severely limits the possible functions we can recover. In this paper, we make use of flexible non-parametric Gaussian Process (GP) priors which allows us to provide significant structure to the learning problem without reducing the possible solution space. The two random variables connected with $f_j(\cdot)$ may then be jointly specified under a GP prior where the covariance, with hyperparameters $\theta$, is evaluated at $G_j$ and $X$ for $F_j^{\mathrm{G}}$ and $F_j^{\mathrm{X}}$ respectively as
\begin{gather}
p\left(\begin{bmatrix} F_j^{\mathrm{X}} \\ F_j^{\mathrm{G}}\end{bmatrix}  \middle| \, G_j, X_j, \theta_j \right) \sim 
\mathcal{N}\left(\mathbf{0}, 
\begin{bmatrix} k_{\theta_j}(X, X) & k_{\theta_j}(X, G_j) \\ k_{\theta_j}(G_j, X) & k_{\theta_j}(G_j, G_j)   \end{bmatrix} 
        \right).%\nonumber
\end{gather}
%\phantom{foo}

\paragraph{Warping functions} We encode our preference for smooth warping functions $g_j(\cdot)$ by making $p(G_j \mid X)$ a GP prior with a smooth kernel function. We can ensure monotonicity by an appropriate parametrisation of the $G_j$ using an auxiliary input. Without loss of generality, these are constrained to be monotonic in the range $[-1,1]$ using a set of auxiliary variables $U_j \in \mathbb{R}^{N}$ such that
\begin{align}
[G_j]_n &:= 2 \,\sum_{k=1}^n\!\left[ \,\mathrm{softmax}\!\left( U_j \right) \right]_{k} - 1\ .
\label{eq:warps}
\end{align}
Importantly, all warping functions are continuous and generative which means we are able to resample the data. Therefore, we write that $p(G_j \mid X, \omega) \sim \mathcal{N}(0, k_{\omega_j}(X, X))$ with hyperparameters $\omega_j$. An alternative to our parameterisation is GPs with monotonicity information~\cite{Riihimaki:2010}, however, this approach does not guarantee that the posterior predictive is monotonic.

\paragraph{Model over sequences} We would like a constraint that aligns similar sequences to each other while keeping dissimilar sequences apart without us specifying which sequences belong together.  We consider using dimensionality reduction as a means of preserving similarities in the prediction space as well as imposing the preference for dissimilar data points to be placed far apart in the latent space. In particular, we propose to use a GP-LVM that places independent GPs over the data features and optimises the locations of the low-dimensional latent points that correspond to each sequence. 

To this end, we let the random variable $Z_j \in \mathbb{R}^{Q}$ be the embedded manifold location of the sequence. The random variable $H_j$ denotes the output of the mapping function evaluated at $Z_j$ such that $H_j \sim h(Z_j)$. To ease notation, we use bold symbols to denote the concatenation across $J$ such that, for example, $\mathbf{Z} = [Z_1, \dots, Z_J]$. We encode the preference for a smooth mapping by placing a GP prior over the mapping $h(\cdot)$ so that we have $p(\mathbf{H} \mid \mathbf{Z}, \psi) \sim \mathcal{N}(0, k_{\psi}(\mathbf{Z}, \mathbf{Z}))$ where $\psi$ are the hyperparameters of the covariance kernel. The pseudo-observations are modelled by the GPLVM by adding independent Gaussian noise to $\mathbf{H}$. Next we consider the joint distribution of the model to derive an objective which simultaneously ensures that (i) the observed data $\mathbf{Y}$ is fitted well by the corresponding GPs $f$ at the warped locations, (ii) the pseudo observations are fitted well by the corresponding GPs $f$ at the fixed sampling locations, (iii) the pseudo observations are such that they exhibit a simple structure that is captured by the latent variable model. 

\paragraph{Joint distribution} The joint distribution (ignoring the hyperparameters and noise terms for clarity) decomposes as
\begin{multline}
p(\mathbf{S},  \mathbf{Y},\mathbf{F}^{\mathrm{X}}, \mathbf{F}^{\mathrm{G}}, \mathbf{G}, \mathbf{H}, \mathbf{Z} | \mathbf{X}) = p(\mathbf{Y} | \mathbf{F}^{\mathrm{G}} ) \, p(\mathbf{S} | \mathbf{H}, \mathbf{F}^{\mathrm{X}}) \\ p(\mathbf{H} | \mathbf{Z}) \, p(\mathbf{F}^{\mathrm{X}}, \mathbf{F}^{\mathrm{G}} | \mathbf{G}, \mathbf{X}) \, p( \mathbf{G} | \mathbf{X}) \, p(\mathbf{Z}) \ . \label{eqn:joint}
\end{multline}
The terms $p(\mathbf{H} | \mathbf{Z})$, $p(\mathbf{F}^{\mathrm{X}}, \mathbf{F}^{\mathrm{G}} | \mathbf{G}, \mathbf{X})$ and $p( \mathbf{G} | \mathbf{X})$ are the GP priors defined previously and $p(\mathbf{Z}) \sim \mathcal{N}(\mathbf{0}, I)$ is the latent prior . We note that $p(\mathbf{F}^{\mathrm{X}}, \mathbf{F}^{\mathrm{G}} | \mathbf{G}, X)$ and $p( \mathbf{G} | X)$ factorise fully over $J$. 

\paragraph{Likelihood terms} The likelihood of the observations under i.i.d. Gaussian noise with precision $\beta_j$ is $p(\mathbf{Y} | \mathbf{F}^{\mathrm{G}} ) = \prod_j p(Y_j | F_j^{\mathrm{G}} ) = \prod_j \mathcal{N}(Y_j | F_j^{\mathrm{G}}, \beta_j^{-1}I)$. The likelihood of the pseudo-observations is more involved since it encodes the relationship of Eq.~\eqref{eqn:such_that}. We define the likelihood $p(\mathbf{S} \mid \mathbf{H}, \mathbf{F}^{\mathrm{X}})$ as an equal mixture:
\begin{gather}
p(\mathbf{S} \mid \mathbf{H}, \mathbf{F}^{\mathrm{X}}) =
\frac{1}{2} \left( \prod_n \mathcal{N}(S_n |\mathbf{H}_n, \gamma^{-1} I_J) + \prod_j \mathcal{N}(S_j | \mathbf{F}_j^X, \beta_j^{-1} I_N) \right) \label{eq:likelihood}
\end{gather} where $S_j$ refers to the rows and $S_n$ refers to the columns of $\mathbf{S}$. 

This explicitly encourages the two components (the one over $h$ and the one over $f$) to coincide so that the pseudo-observations are explained by both components of the model simultaneously. In order to find the maximum likelihood solution, we use the fact that $\log(\rfrac{1}{2} \: a + \rfrac{1}{2} \:b) \geq \rfrac{1}{2}\log(a) + \rfrac{1}{2}\log(b)$, and we maximise the lower bound on the log-likelihood to find the parameters of the two models, the latent variables, and the pseudo-observations.

\paragraph{Approximations} The integrals over $\mathbf{H}$ and $\mathbf{F}^X$ are regular GP marginalisations, which can be computed in closed-form. The integral corresponding to $\mathbf{F}^{\mathrm{G}}$ includes a composition of GPs, ($f \circ g$), which does not have a closed-form solution. Following \cite{Lawrence:2007hierarchical}, we approximate this integral using a point estimate, and since $\mathbf{G}$ is directly optimised, it allows us to use the monotonic parametrisation of Eq.~\eqref{eq:warps} without the need to integrate over the corresponding parameters.

\paragraph{Learning}We place priors over the hyperparameters $\{\gamma, \psi, \thetaj, \betaj\}$ as log-Normal distributions with zero mean and unit variance. We also place an additional prior on the raw sample points $U_j$ to encourage smooth warps, and improve training as,
\begin{equation}
\log \, \p{\{U_j\}} = \displaystyle\sum_{j=1}^{J} \, \log \,\NormalDistrib{U_j\mid \mathbf{0}, I_{N}}.
\end{equation} We optimise the following marginal log-likelihood (excluding the terms corresponding to the priors on the hyperparameters): $\log p(\mathbf{S}, \mathbf{Y}, \mathbf{Z} \mid X)  = \log p(\mathbf{S}, \mathbf{Y} \mid X) + \log p(\mathbf{S} \mid  \mathbf{Z}) + \log p(\mathbf{Z})$ w.r.t. the pseudo observations $\mathbf{S}$, the latent variables $\mathbf{Z}$ and the hyperparameters of the model to obtain the MAP estimates. 

\paragraph{Implementation}We implement our model using the TensorFlow~\cite{Tensorflow} framework and minimise the negative marginal log-likelihood objective using the Adam optimizer~\cite{AdamOpt}. By default, we used standard squared exponential covariance functions for all the Gaussian process priors.
In some of the experiments, different covariance functions were used when the data or warping functions were less smooth (\eg the Mat\'{e}rn covariance).
The complexity of our method is limited by the inversion of the covariance matrices and therefore scales with $\mathcal{O}(J N^3 + J^3)$. % $\mathcal{O}\!\left((2N)^3\right)$. 
However, there are standard sparse approaches available to scale to longer sequences. We also implemented the sparse variational method of Titsias~\cite{Titsias:2009} which reduces the complexity to $\mathcal{O}\!\left(J\,NM^2 + J^3\right)$, where $M$ is a specified number of inducing points for the sparse approximation. This method performed well for $M$ an order of magnitude smaller than the full $N$. We note that the use of a sparse approximation fits naturally with the rest of our model as it increases the smoothness of the observations, which may simplify the alignment task.

\subsection{Comparison of Variants of our Model} 
\label{sec:param}
Our proposed model is fully non-parametric and models both the warping and the generating functions at the same time. Methods that rely on the standard $\mathbb{L}^2$ metric in the input space are ill-posed and thus require a regularisation term. This leads to an optimisation problem that suffers from poor local minima and relies on the use of a coarse-to-fine approach. In order to highlight the limitations of using the standard $\mathbb{L}^2$ metric in the input space, we describe a model that performs a parametric re-sampling of the data which corresponds to removing our model of the warpings but retaining a model of the data. In effect we take a traditional pairwise minimisation approach but include a probabilistic model of the data which has the effect of regularising the optimisation problem.
 
\paragraph{Parametric warps}We use a parametric re-sampling function $\tilde{g}^{(j)}(\cdot)$ similar to \cite{Zhou:2016} consisting of $K$ monotonically increasing basis functions. For each input sequence $Y_j$, we learn a set of weights $\mathbf{w}^{(j)}\in\mathbb{P}^{K}$. By enforcing that the weights lie on the surface of the $k^{\text{th}}$ order probability simplex $\mathbb{P}$ the resulting function is guaranteed to be monotonic. The task is now to find the set of weights $\{\mathbf{w}_k^{j}\}_{k=1}^K$ such that resampling the data according to the warping functions results in the aligned sequences. As we do not have access to $\mathbf{S}$, we use the same latent variable model as previously and refer to this model as \emph{GP-LVM+basis}. The model can be learned using gradient descent. The parametric model described above, as well as some previous approaches, rely on hand-picked basis functions to define the warps. This results in poor accuracy when the set of basis functions is small and in high computational complexity when the set is large.

\paragraph{Energy alignment}We demonstrate the efficacy of using the alignment GP-LVM to perform simultaneous clustering and alignment by replacing it with an energy minimisation objective that is similar to the previous literature, \eg~\cite{Kurtek:2011}. The latent variable model part of the objective is replaced with an energy minimisation term between each of the $S_j$ and the mean of all the sequences $\{S_j\}_{j=1}^J$. In \S~\ref{sec:experiments} we show the results of this method with the GP warping functions (\emph{energy+GP}) and with the basis function warpings as described above (\emph{energy+basis}).
% -----------------------------------------------------------------------------------------

%% file: includes/overview_figure.tex
% !TEX root = ../alignment_aistats.tex

\begin{figure*}[h!]
\centering
\begin{subfigure}[h]{\textwidth}
            \includegraphics[width=0.19\textwidth, height=1.8cm]{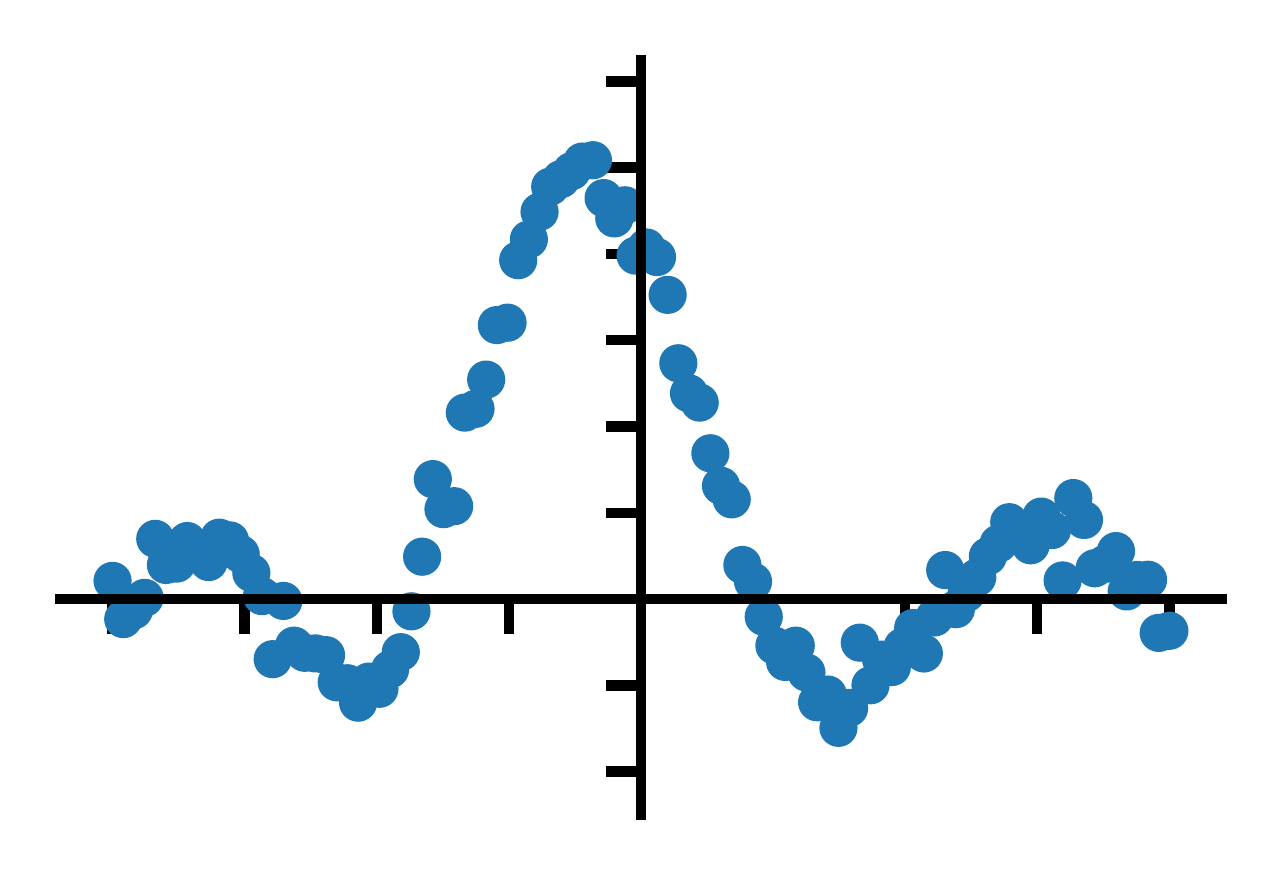}\hfill%
            \includegraphics[width=0.19\textwidth, height=1.8cm]{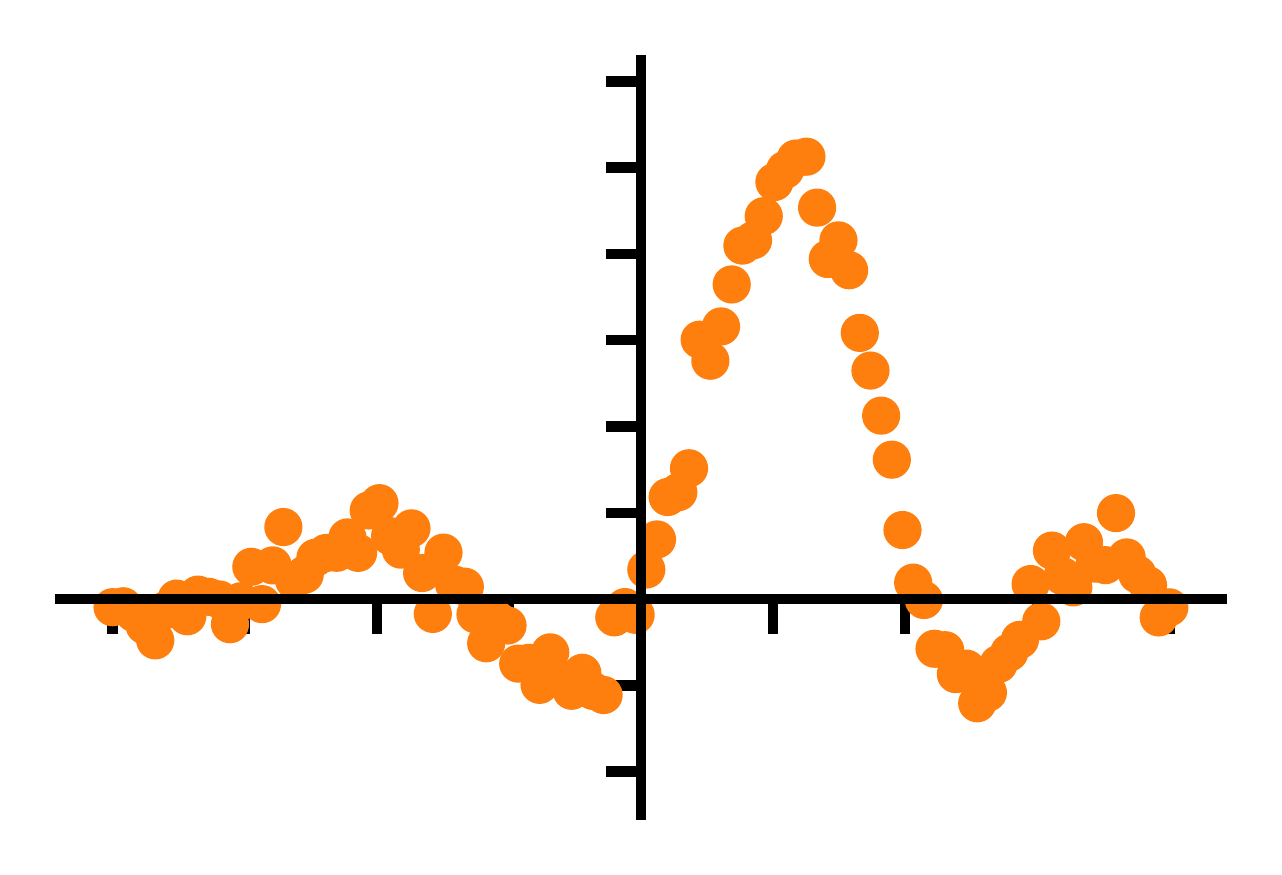}\hfill%
            \includegraphics[width=0.19\textwidth, height=1.8cm]{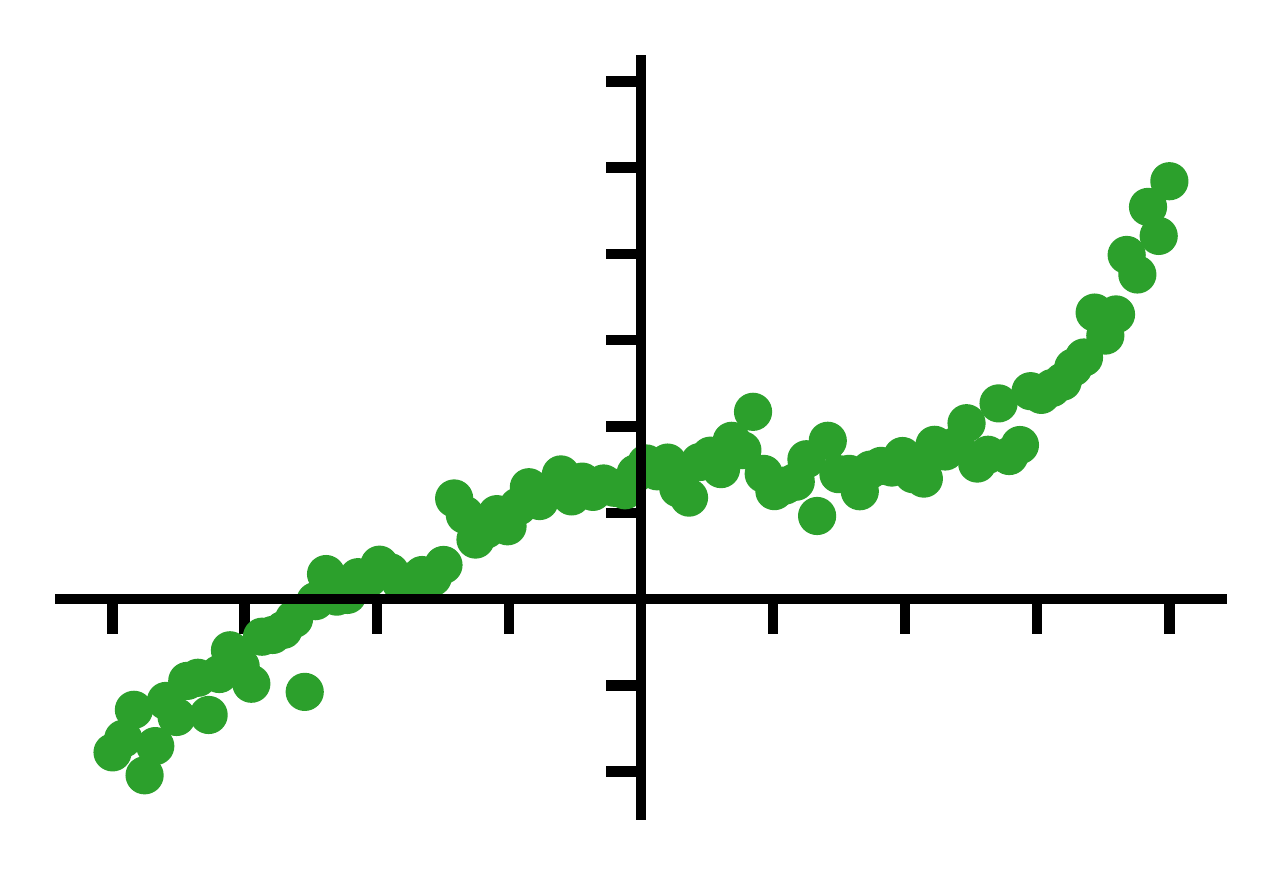}\hfill%
            \includegraphics[width=0.19\textwidth, height=1.8cm]{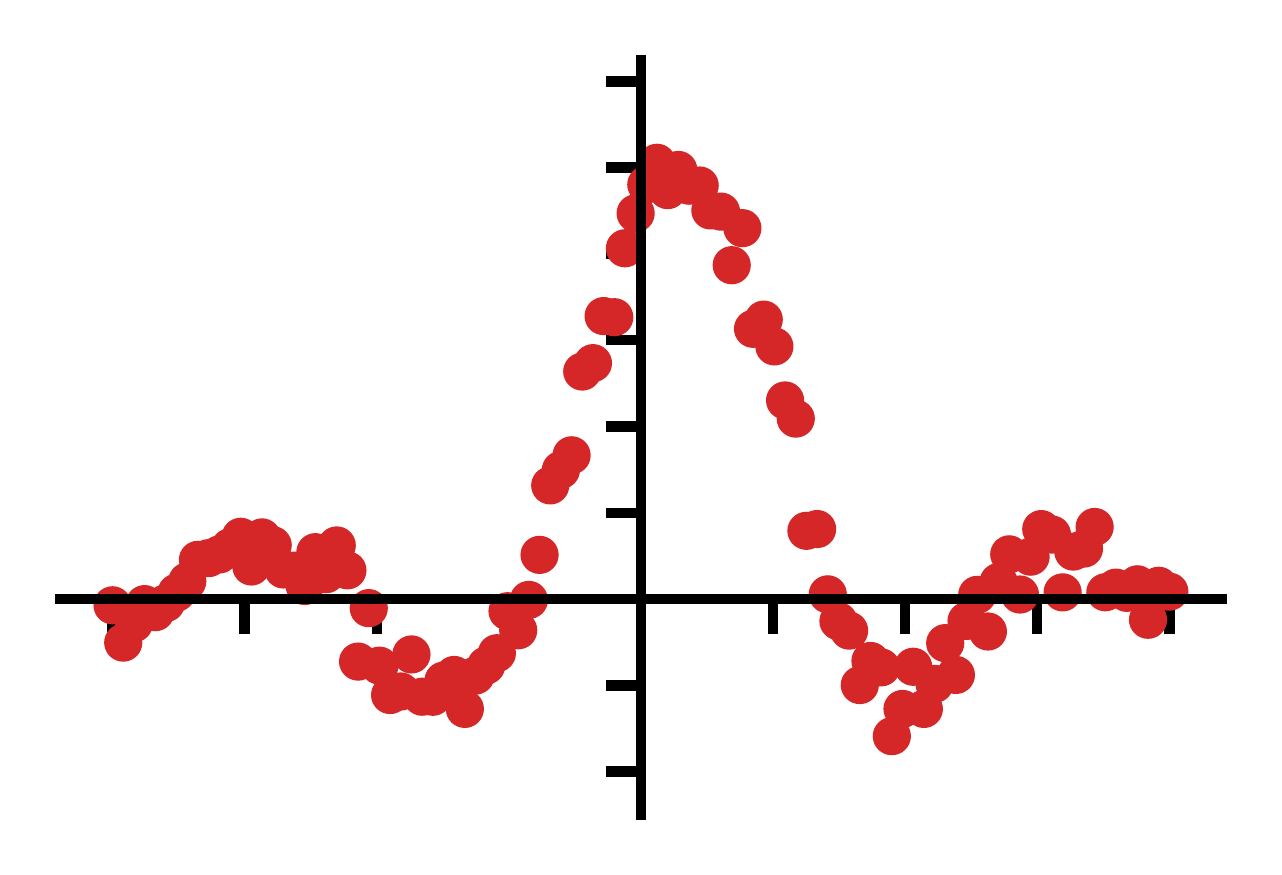}\hfill%
            \includegraphics[width=0.19\textwidth, height=1.8cm]{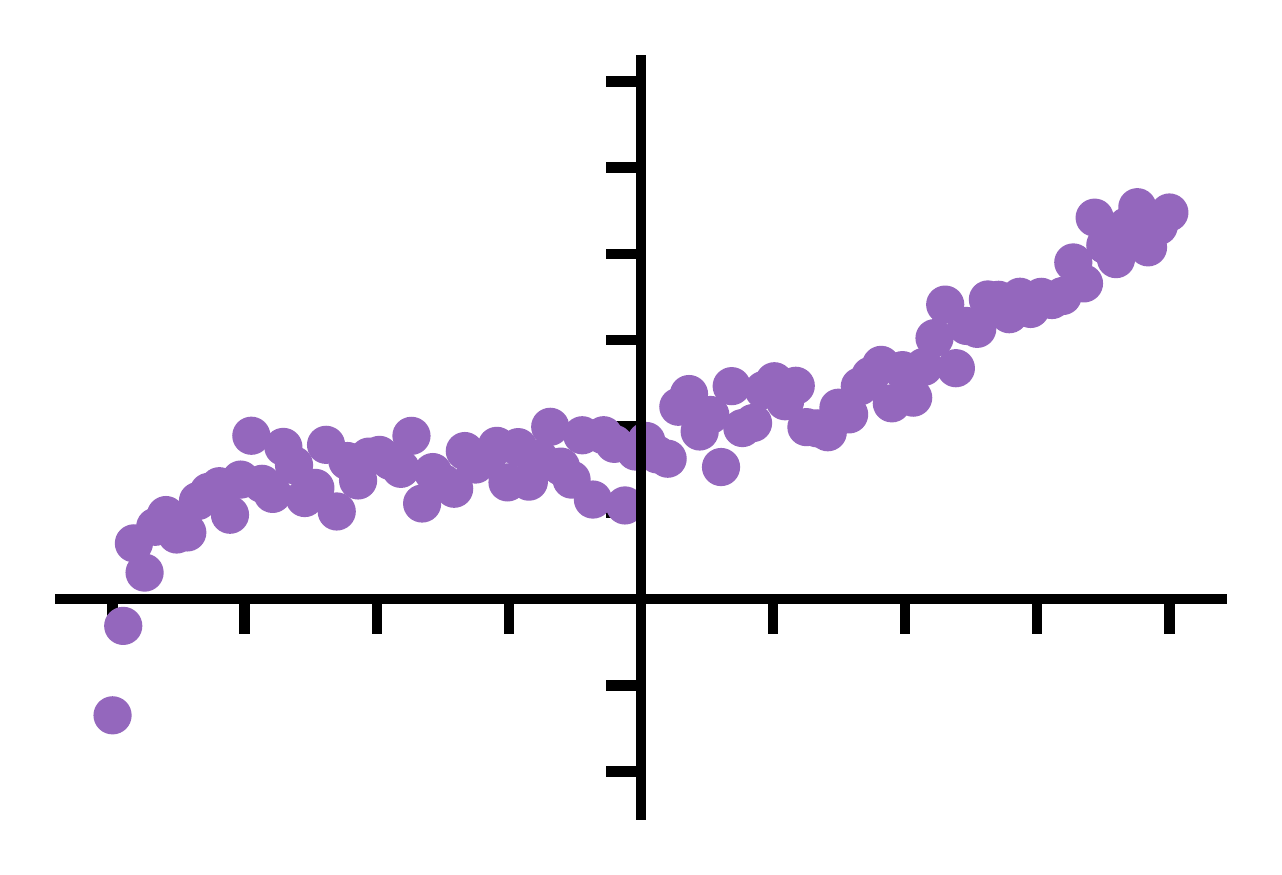}%
	\caption{The observed data $Y_j$, $j = 1 \dots 5$, are noisy and time warped versions of the unknown true sequences.}\label{fig:overview_obs}
    \end{subfigure}
\begin{subfigure}[h]{\textwidth}
            \includegraphics[width=0.19\textwidth, height=1.8cm]{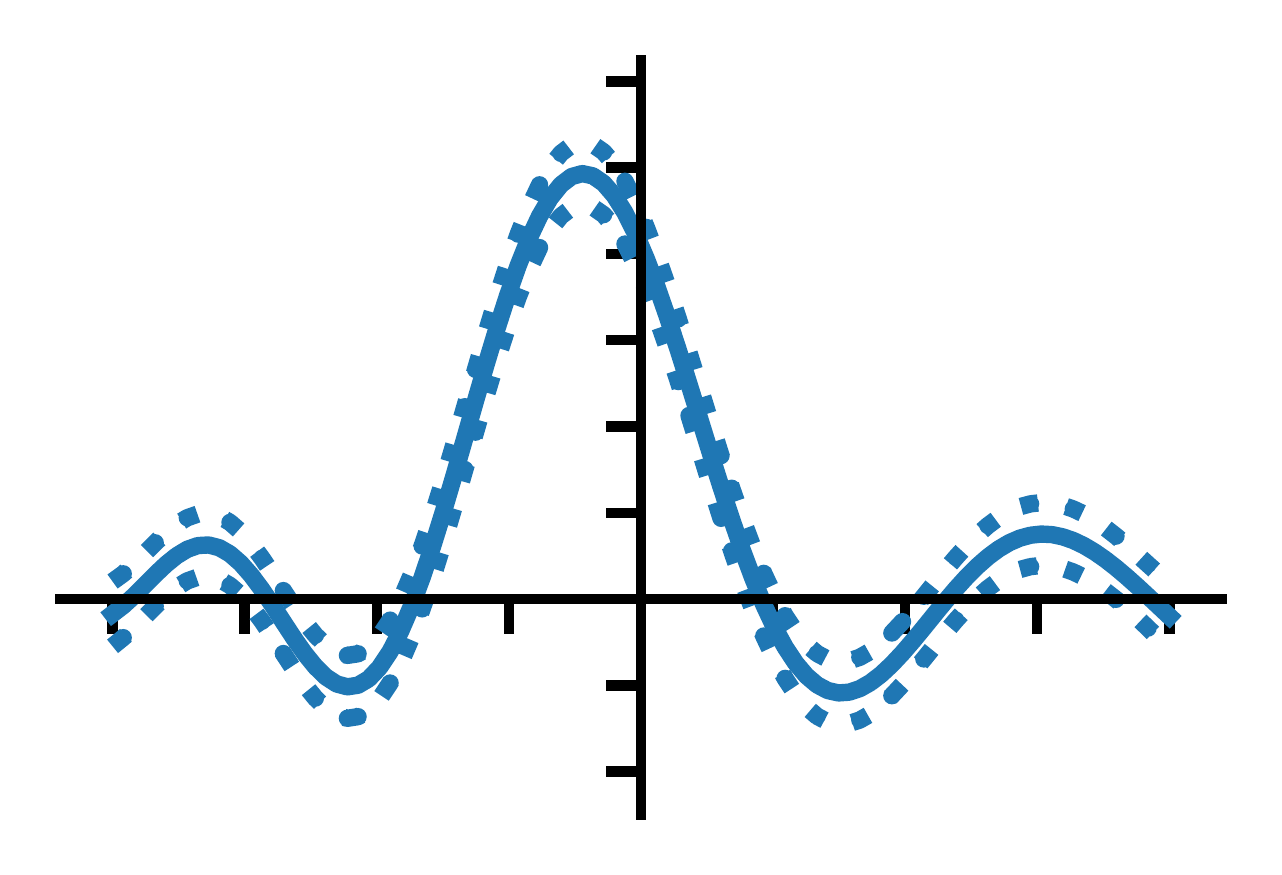}\hfill%
            \includegraphics[width=0.19\textwidth, height=1.8cm]{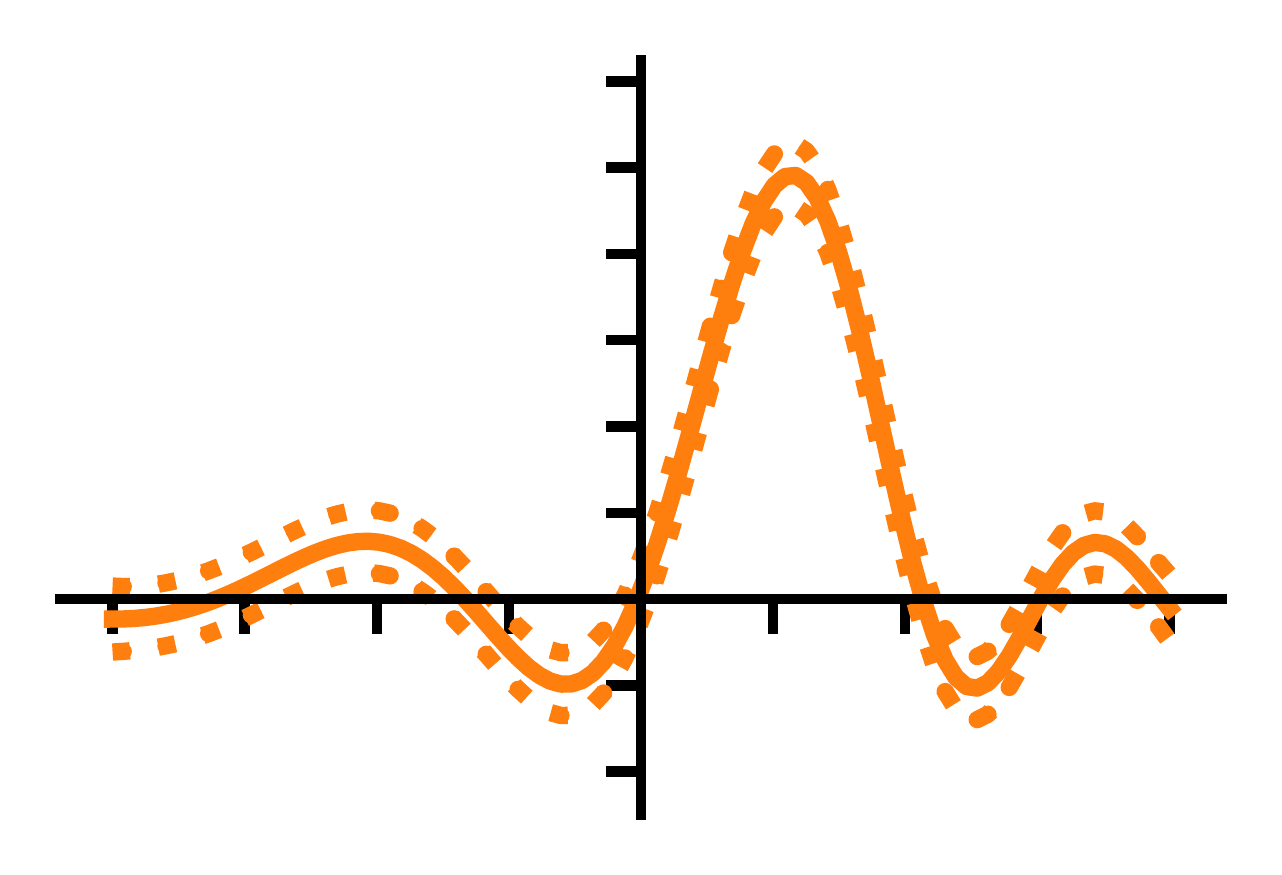}\hfill%
            \includegraphics[width=0.19\textwidth, height=1.8cm]{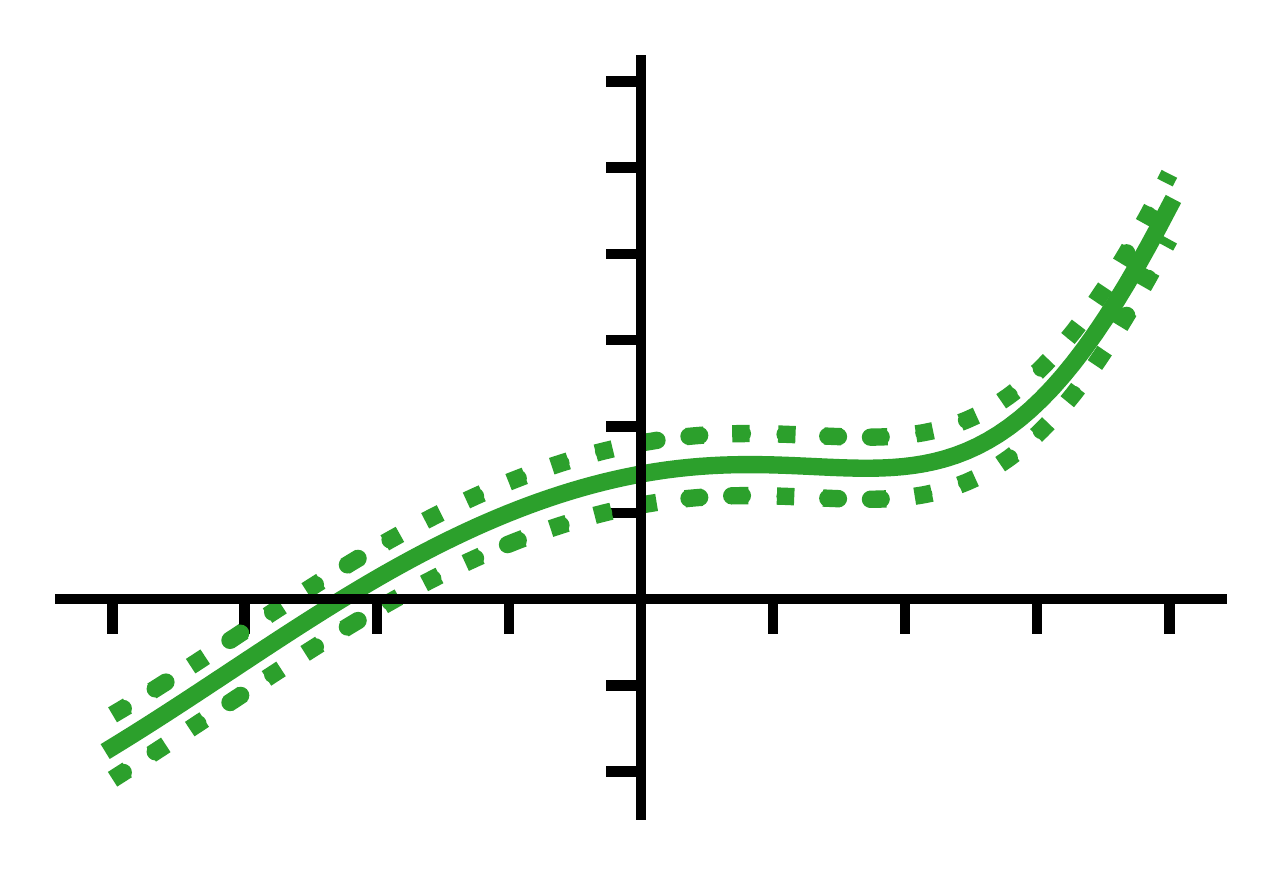}\hfill%
            \includegraphics[width=0.19\textwidth, height=1.8cm]{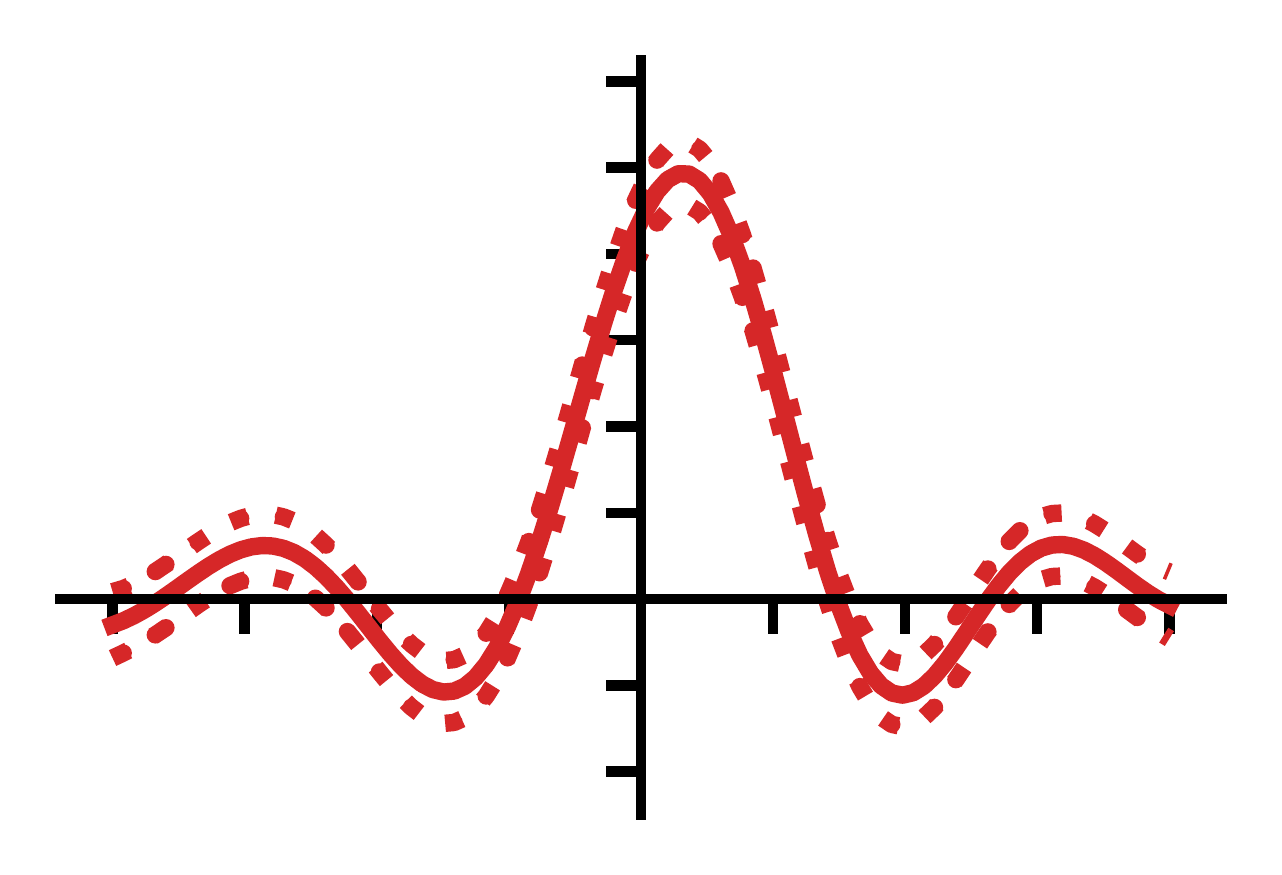}\hfill%
            \includegraphics[width=0.19\textwidth, height=1.8cm]{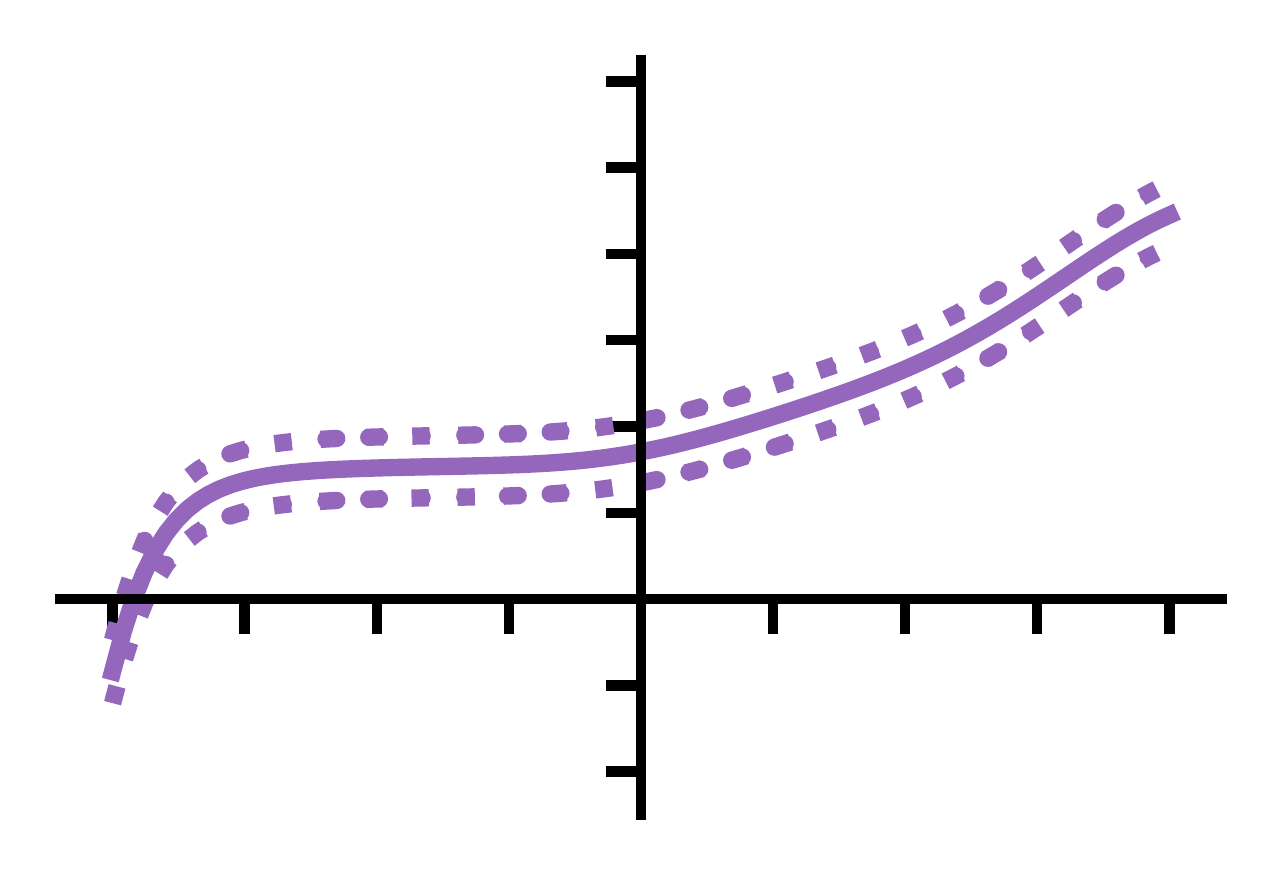}%
	\caption{Continuous functions are fit (with uncertainty) to the noisy, time warped observations (these are $f_j\big(\,g_j(x)\big)$).}\label{fig:overview_fit}
    \end{subfigure}
\begin{subfigure}[h]{\textwidth}
            \includegraphics[width=0.19\textwidth, height=1.8cm]{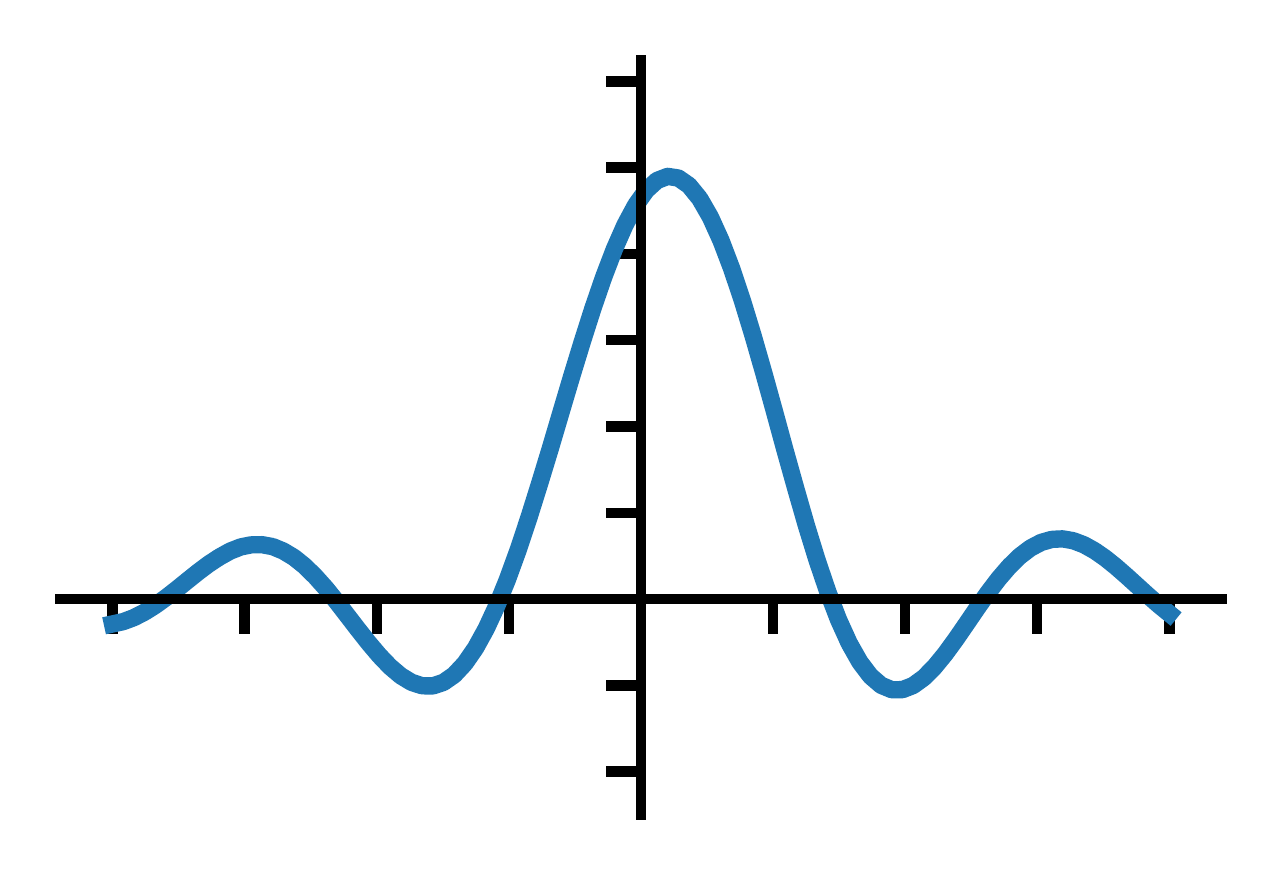}\hfill%
            \includegraphics[width=0.19\textwidth, height=1.8cm]{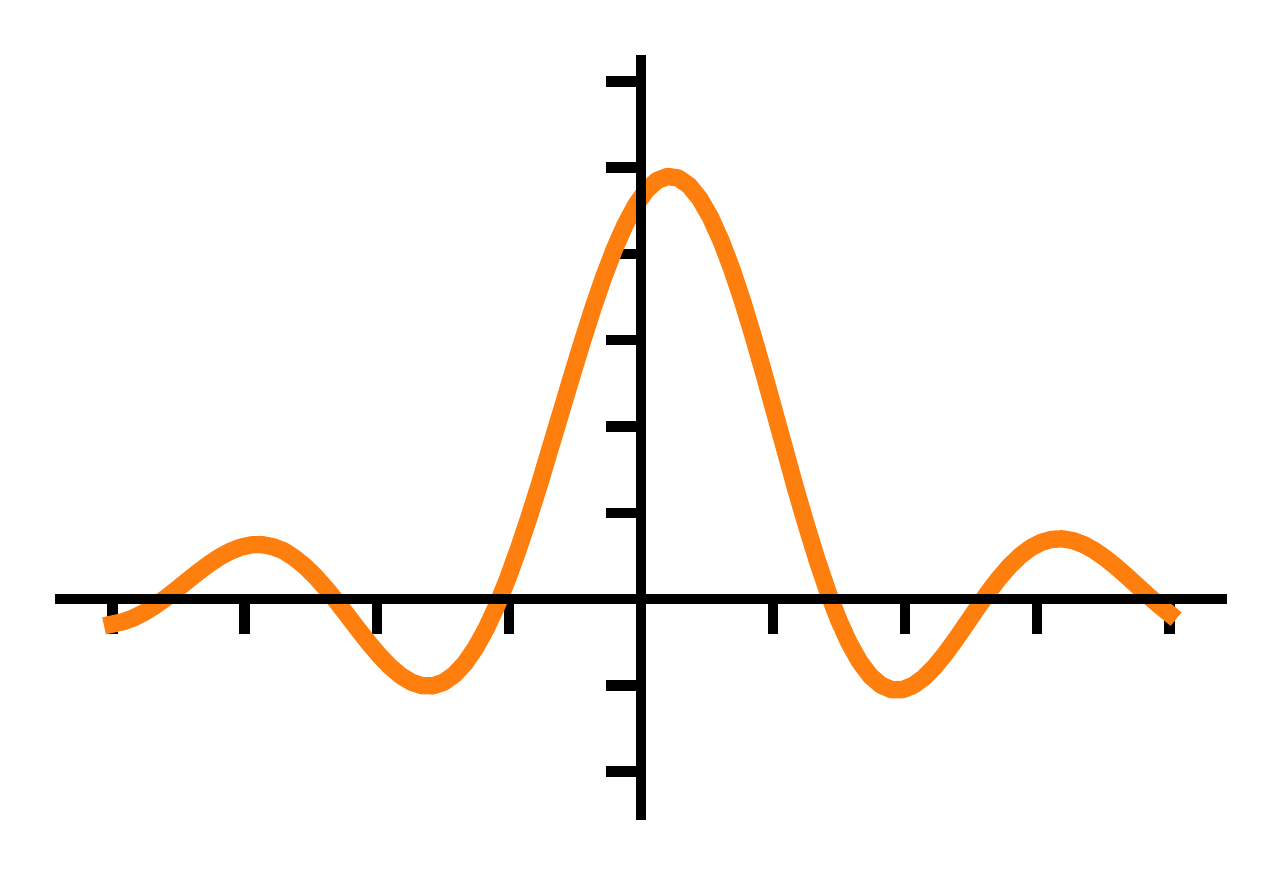}\hfill%
            \includegraphics[width=0.19\textwidth, height=1.8cm]{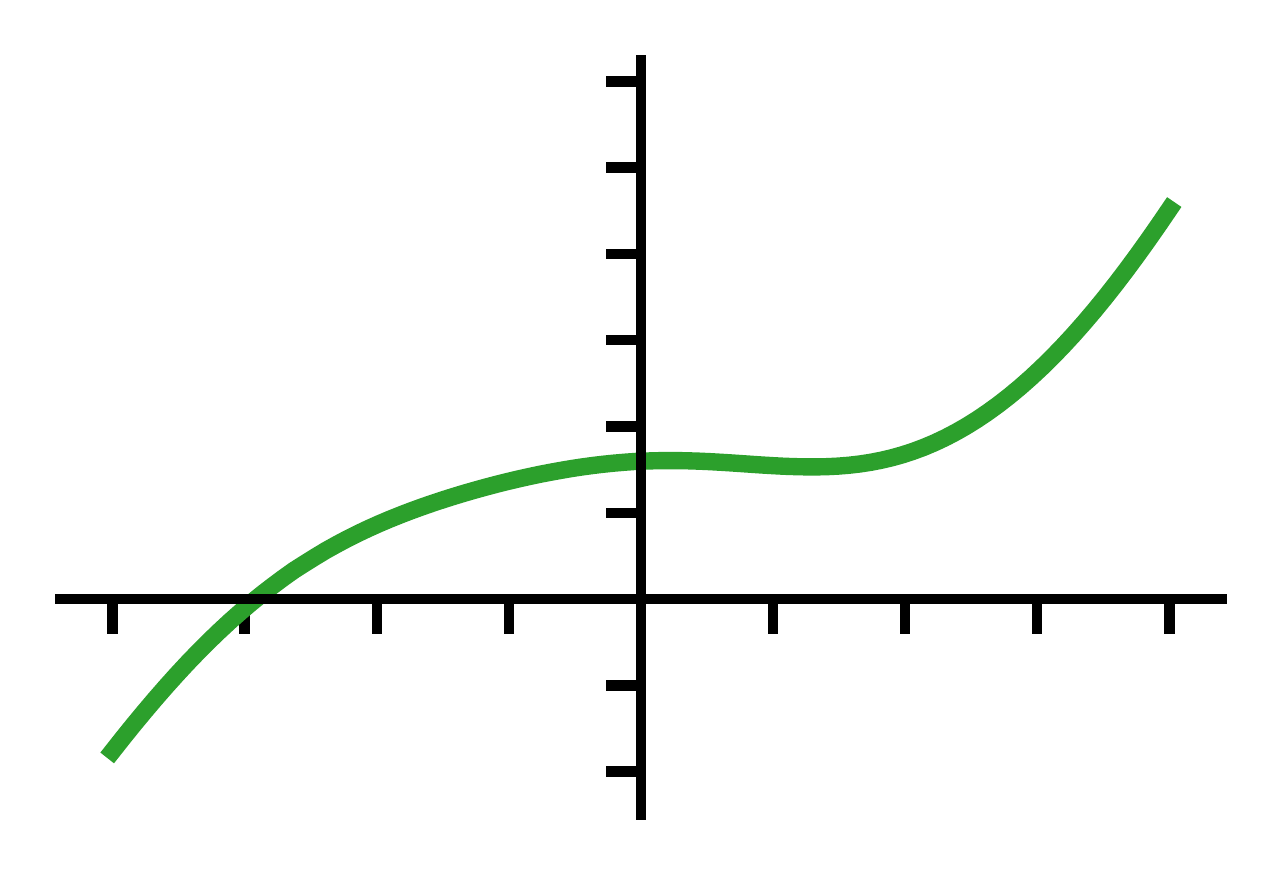}\hfill%
            \includegraphics[width=0.19\textwidth, height=1.8cm]{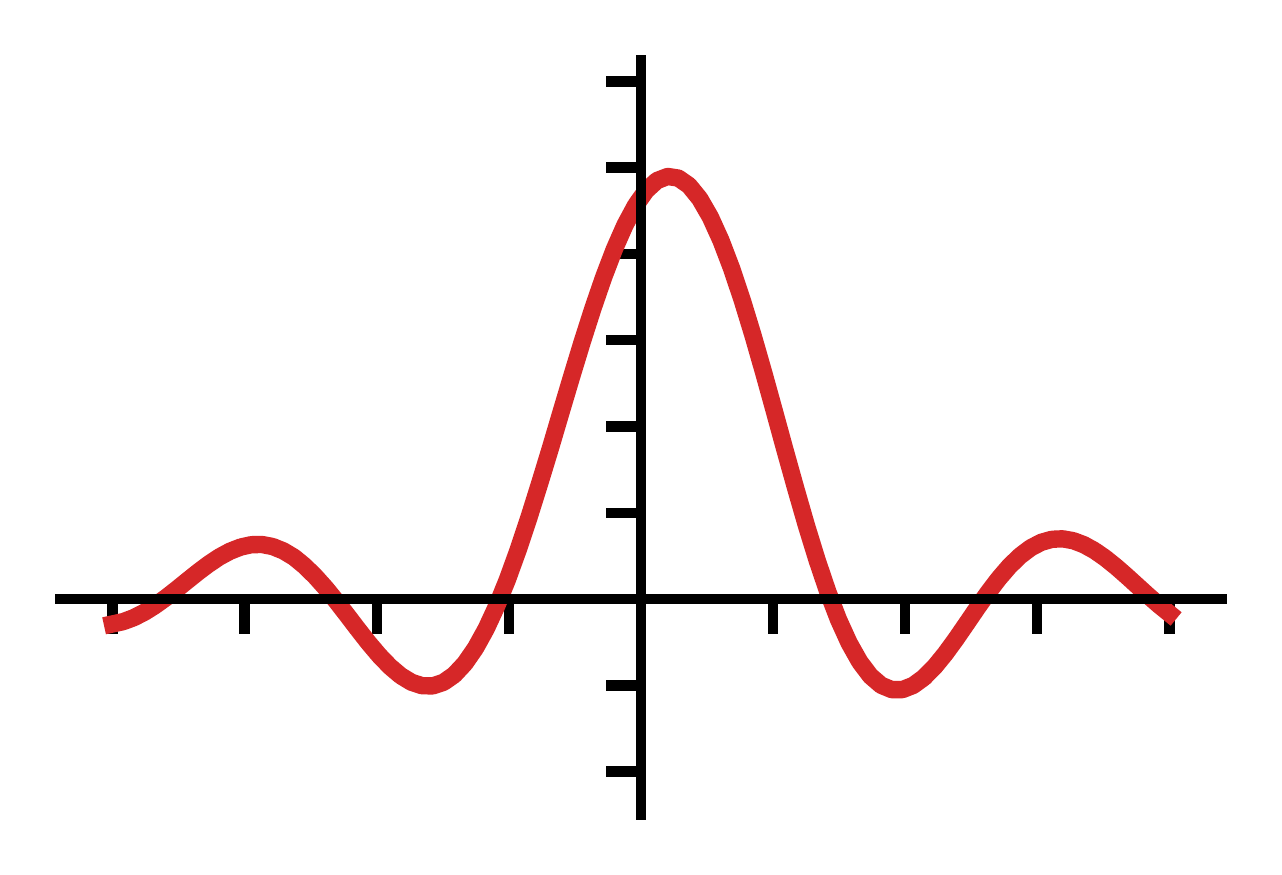}\hfill%
            \includegraphics[width=0.19\textwidth, height=1.8cm]{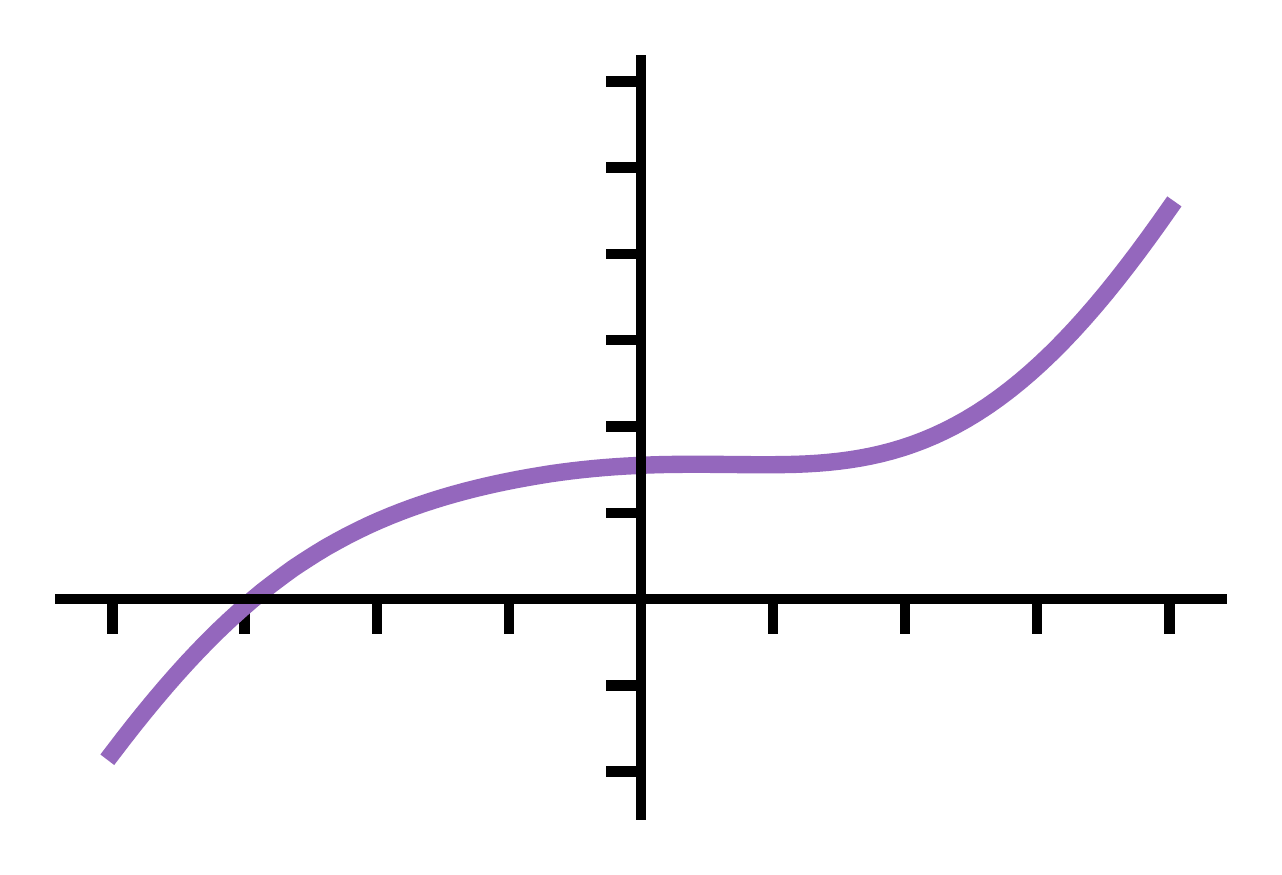}\\%
            \includegraphics[width=0.19\textwidth, height=.8cm]{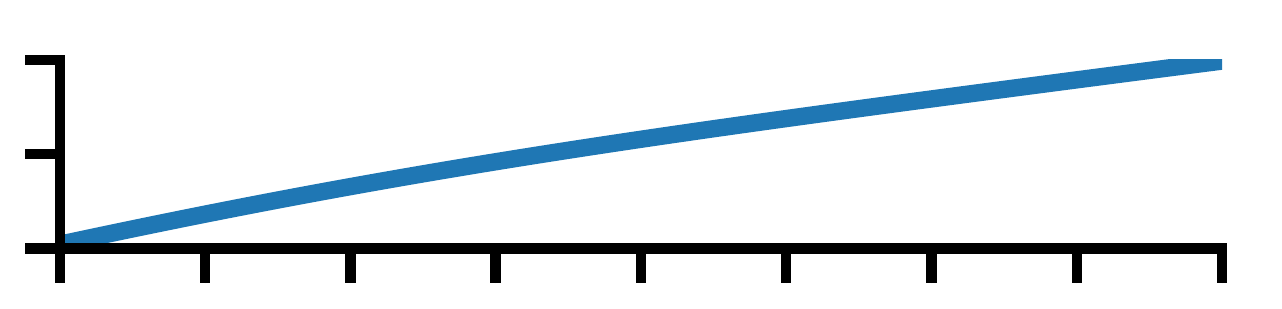}\hfill%
            \includegraphics[width=0.19\textwidth, height=.8cm]{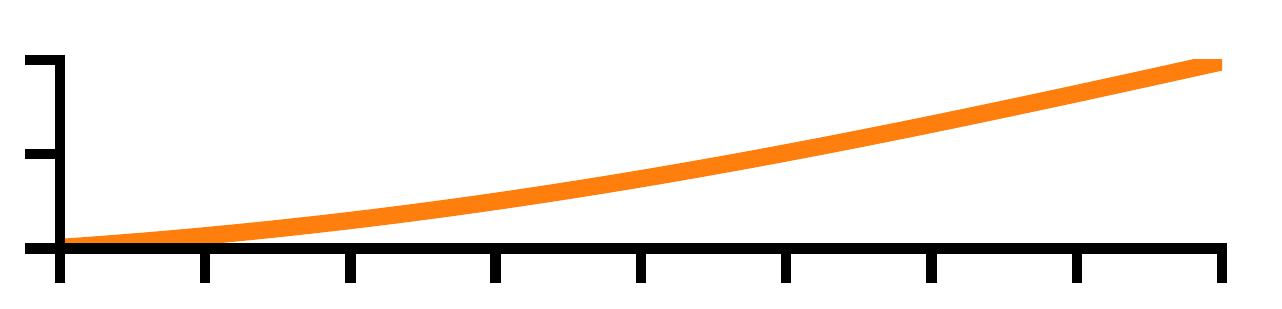}\hfill%
            \includegraphics[width=0.19\textwidth, height=.8cm]{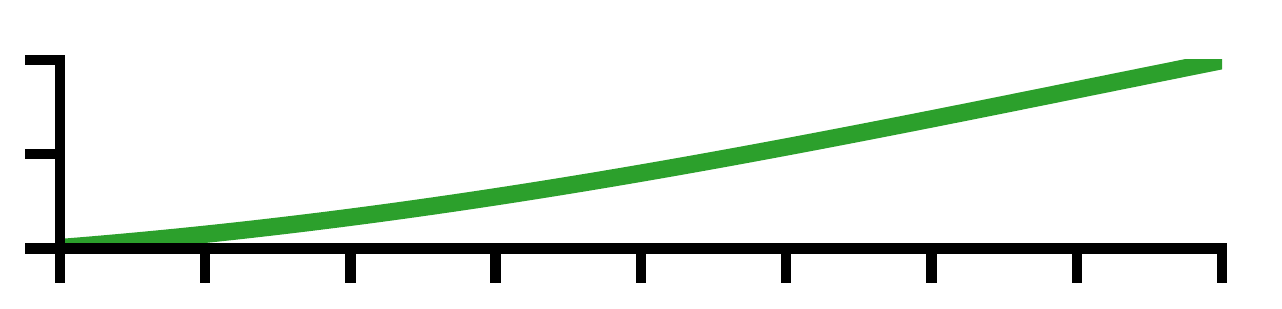}\hfill%
            \includegraphics[width=0.19\textwidth, height=.8cm]{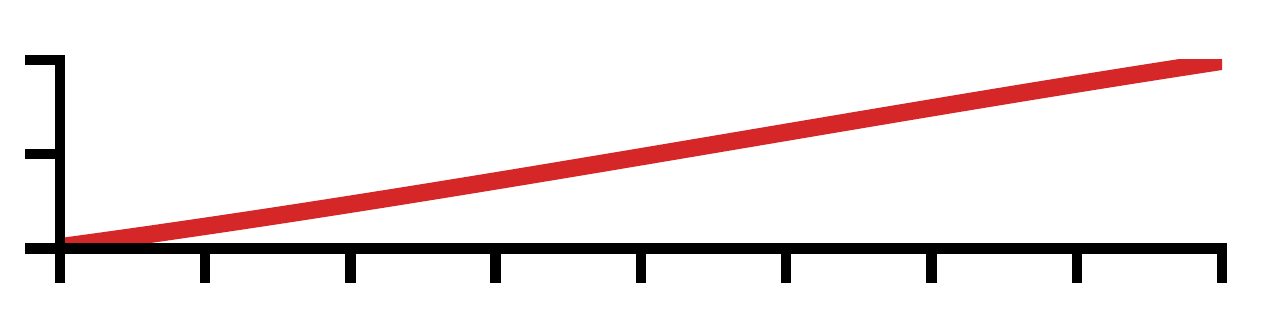}\hfill%
            \includegraphics[width=0.19\textwidth, height=.8cm]{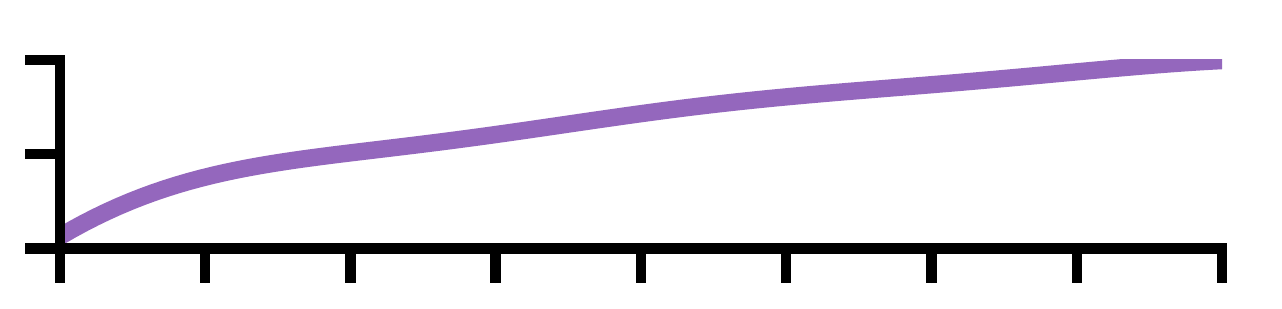}%
	\caption{These are decomposed into the aligned sequences $f_j(x)$ (top row) and the inferred continuous warp $g_j(x)$ (bottom row).}\label{fig:overview_aligned}
\end{subfigure}
        
\begin{minipage}{0.32\textwidth}%
    \begin{subfigure}[h]{\textwidth}\centering
            \includegraphics[width=0.8\textwidth, height=1.2cm]{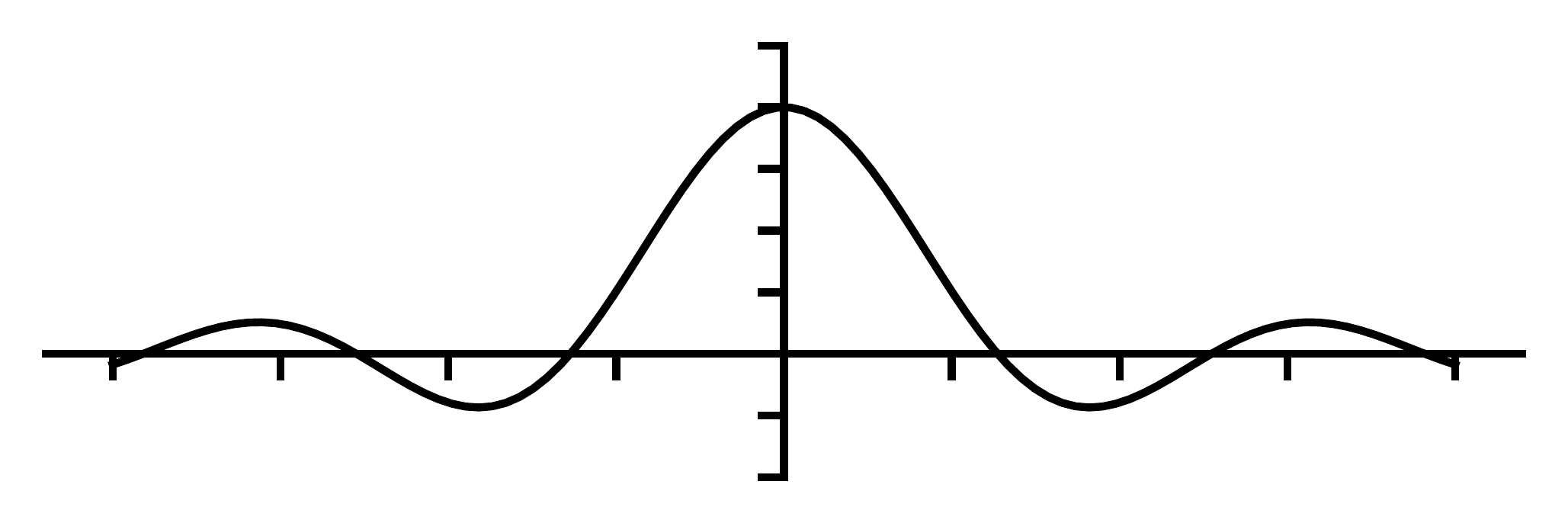}\\%
    		\includegraphics[width=0.8\textwidth, height=1.2cm]{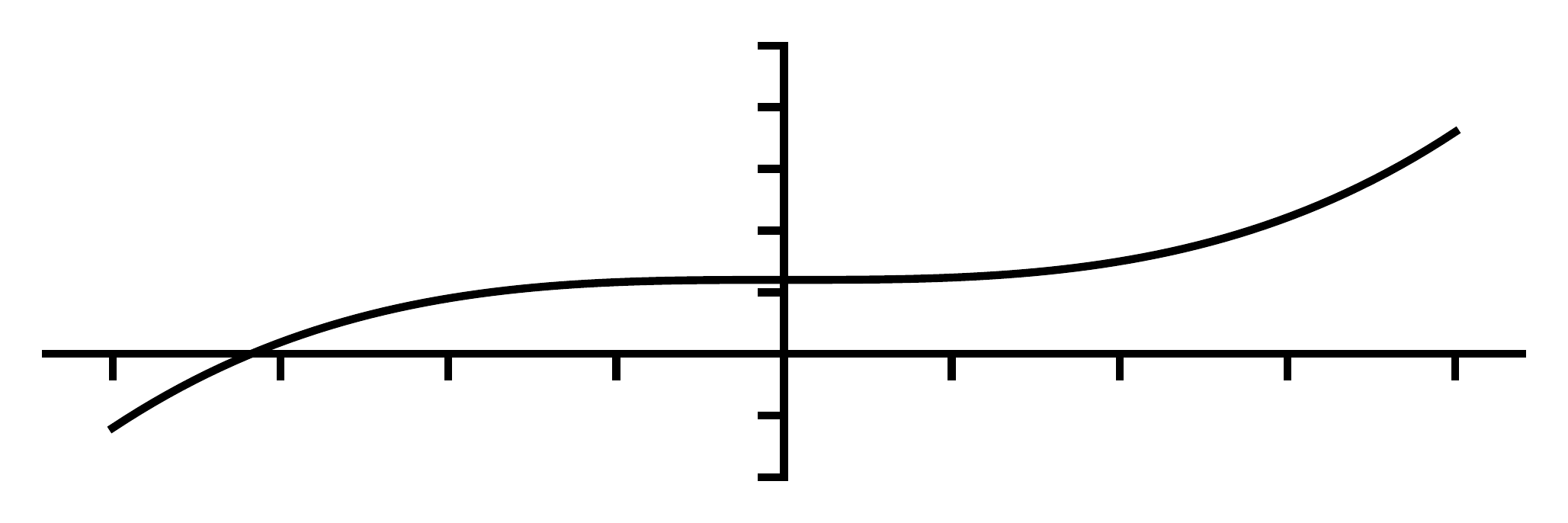}%
    \caption{The set of two true unobserved sequences for the warped observations.}\label{fig:overview_true}
    \end{subfigure}%
\end{minipage}\hfill%
\begin{minipage}{0.32\textwidth}%
    \begin{subfigure}[h]{\textwidth}\centering
             \includegraphics[width=0.8\textwidth, height=2.5cm]{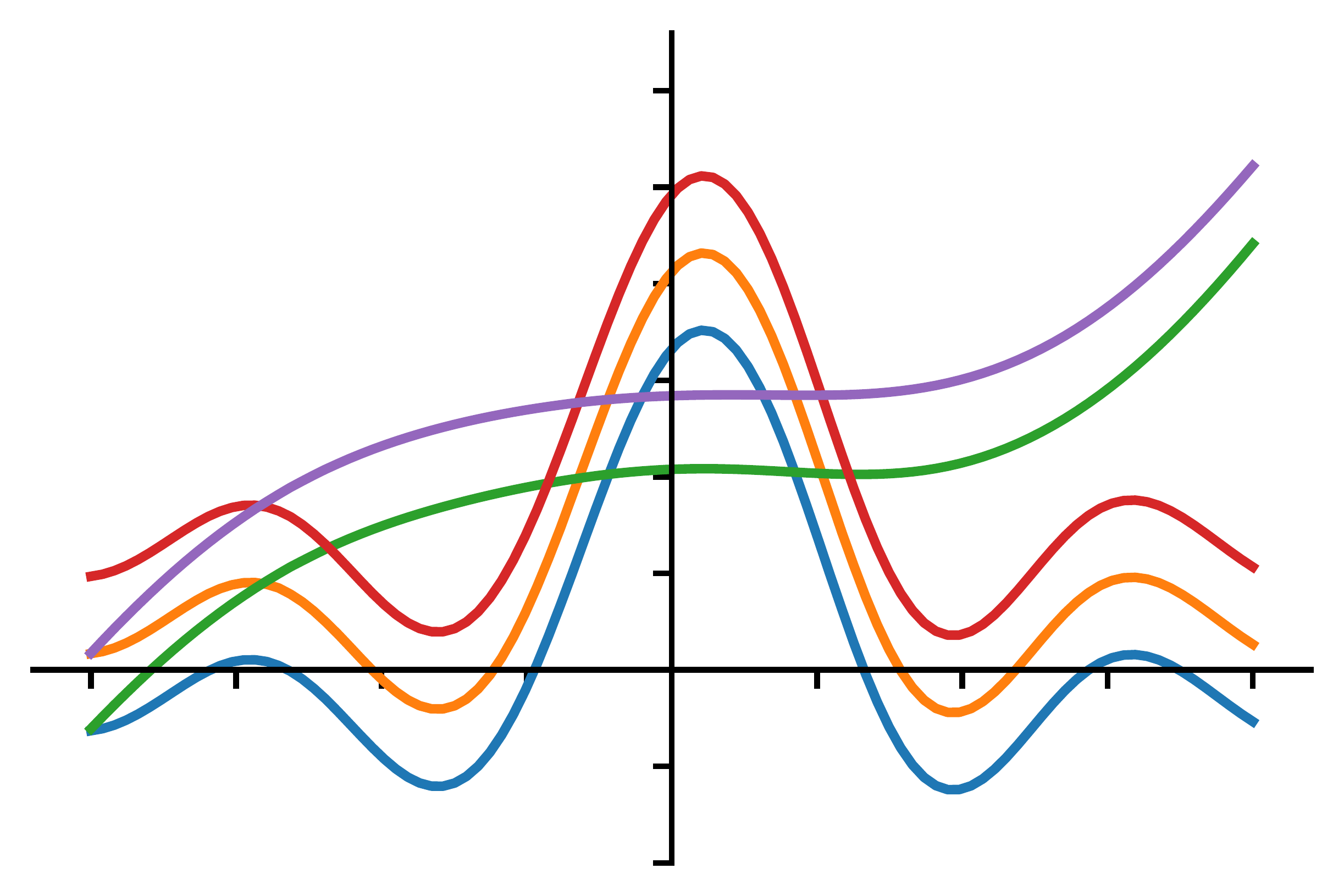}%
    \caption{The inferred aligned results (shown offset in the y-axis).}\label{fig:overview_all_aligned}
    \end{subfigure}
\end{minipage}\hfill%
\begin{minipage}{0.32\textwidth}%
    \begin{subfigure}[h]{\textwidth}\centering
             \includegraphics[width=0.8\textwidth, height=2.5cm]{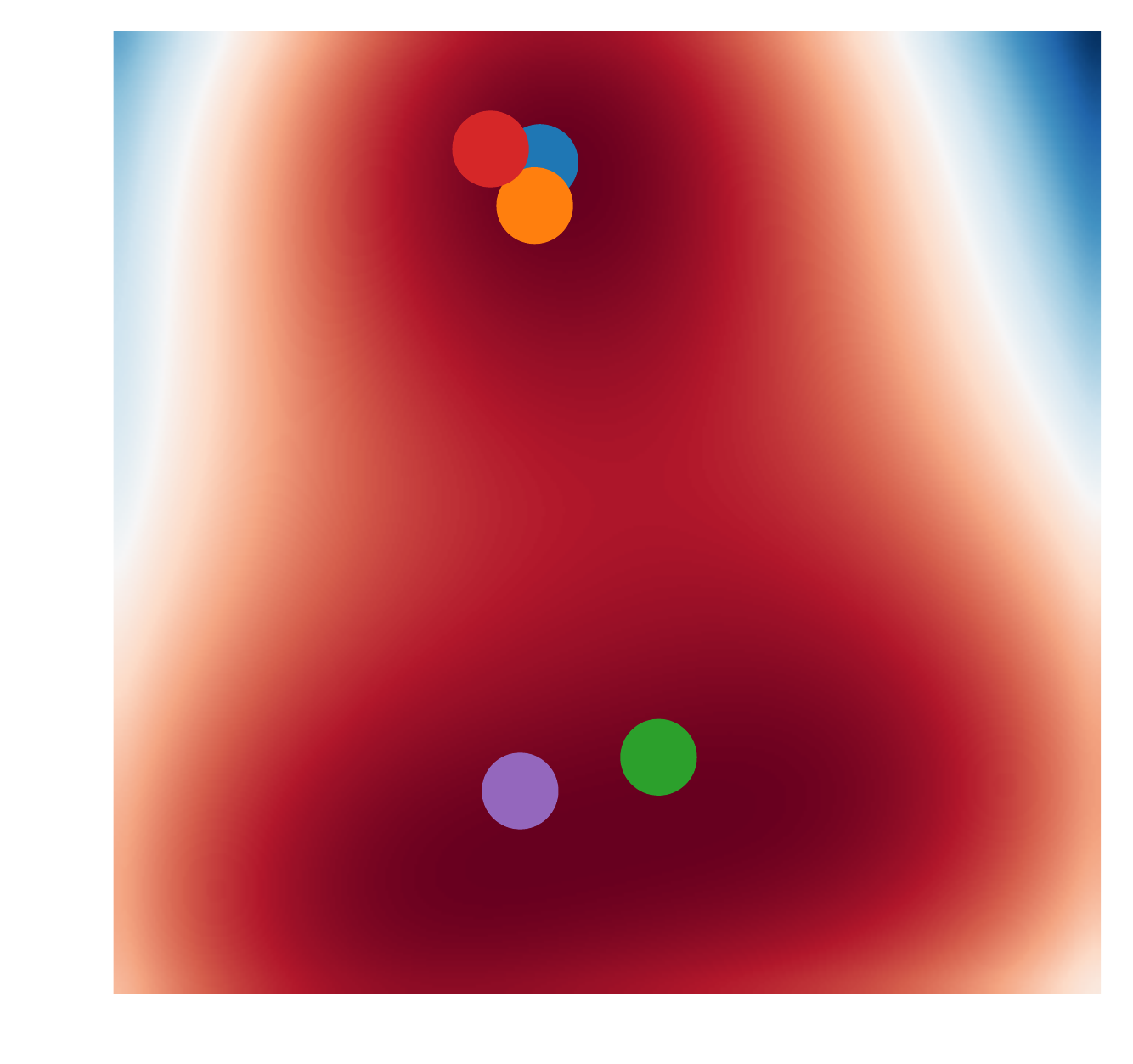}%
    \caption{The inferred generative manifold clusters the two sequence types.}\label{fig:overview_manifold}
    \end{subfigure}
\end{minipage}
        
\caption{Overview of our model on a toy example. We are presented with a set noisy observations (\subref{fig:overview_obs}) that we assume to be time warped versions from a set of true sequences (\subref{fig:overview_true}). We fit continuous functions (\subref{fig:overview_fit}) to the observations and then decompose and cluster them into aligned versions with continuous time warps (\subref{fig:overview_aligned}). This results in a generative model where the aligned sequences (\subref{fig:overview_all_aligned}) are produced from a manifold (\subref{fig:overview_manifold}) that reveals the clustering of the observations into the two fundamental sequences of (\subref{fig:overview_true}).} \label{fig:overview}      
\end{figure*}

%% file: includes/experiments.tex
% !TEX root = ../alignment_aistats.tex
\section{Experiments} \label{sec:experiments}
We now discuss the experimental evaluation of our proposed model. We use standard squared exponential covariance functions for all the GP priors, unless stated otherwise. We show comparisons to current state-of-the-art approaches from data mining and functional data analysis communities using publicly available reference implementations\footnote{See~\cite{CTW:implement} for the implementation of DTW, DDTW, IMW, CTW, GTW, and~\cite{SRVF:implement} for the implementation of SRVF.}. The accuracy is primarily measured in terms of the warping error, \ie~the mean squared error (MSE) between the known true warps and the estimated warps. The alignment error, the MSE between the pairs of aligned sequences, is easily misinterpreted since it is a local measurement. In particular, it does not capture the degenerate cases where the local maxima and minima in the input sequences are shifted to non-corresponding extrema; this is particularly true in datasets with periodic components. Other examples of degenerate behaviour are multiple dimensions collapsing to a single point and warps that rely on translating and rescaling every input in each dimension that leads to over-fitting (an example of this is IMW alignment~\cite{Zhou:2012}). All of these result in high alignment accuracy but produce poor quality results. 

\begin{figure}[t!]
\begin{minipage}{0.49\textwidth}
\centering
    \includegraphics[width=0.92\textwidth]{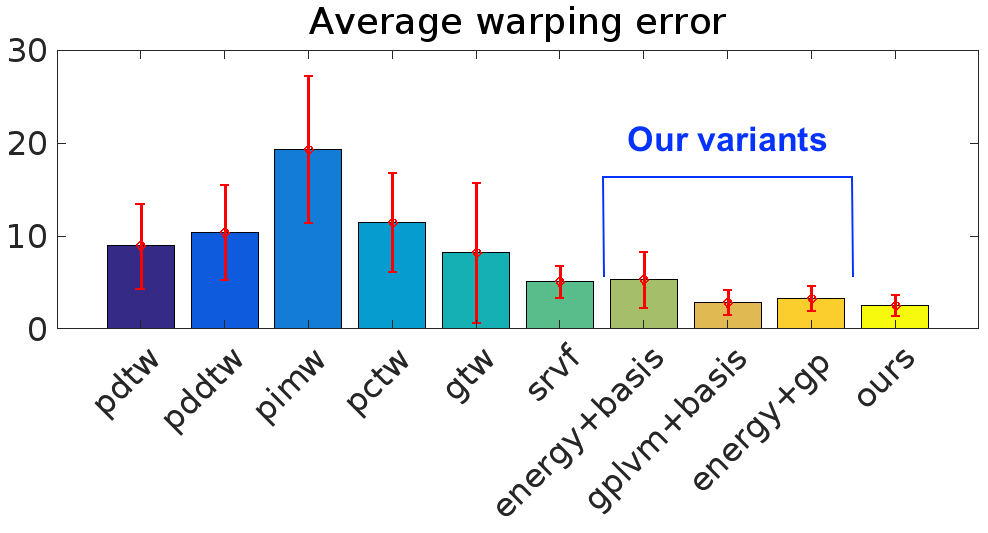}
    \caption{Comparison to state-of-the-art: average error on $25$ datasets proposed by Zhou \etal~\cite{Zhou:2012}. } \label{fig:toys} % See also Table~\protect\ref{table:main_results}.      
%\vspace{0.5cm}
\end{minipage}
\end{figure}

\paragraph{Datasets with quantifiable comparisons} \label{sec:toys}
For this experiment, we use the dataset proposed by Zhou and De la Torre~\cite{Zhou:2012}. It consists of sequences that are generated by temporally transforming latent 2D shapes under known warping transformations that allow quantitative evaluation of the estimated warps. To better assess the quality of the alignments, we run $25$ tests with randomly selected size of the dataset, dimensionality and temporal transformations. Our approach outperforms other methods on these datasets, see Fig.~\ref{fig:toys} and the supplementary material, and produces accurate alignments irrespective of the size of the dataset, dimensionality and structure of the sequences. 
\begin{table}
\begin{minipage}{0.49\textwidth}
% \vskip -.5in
\setlength{\tabcolsep}{4pt}
\centering
\small
\scalebox{0.9}{
\begin{sc}
\begin{tabular}{lcccr}
\hline \\[-1pt]
MSE (SD) & srvf & gp-lvm+basis & \textbf{Ours} \\[1pt]
\hline \\[-1pt]
Alignment & 6.4 ($\pm$1.7) & 8.4 ($\pm$2.7) & \textbf{5.9} ($\pm$1.1) \\[2pt] 
Warping & 30.0 ($\pm$10.4) & \textbf{9.7} ($\pm$4.9) & \textbf{9.7} ($\pm$5.7) \\[1pt]
\hline \\[-12pt]
\end{tabular}
\end{sc}
}%scalebox
%\vskip -2mm
\caption{Quantitative comparison of alignments and warps for the best competing method on dataset with multiple true sequences (alignment and grouping task).}\label{table:quantitative}
%\vskip -0.1in
\end{minipage}
\end{table}
The variant of our method that uses parametric warps (\emph{gplvm+basis}) performs competitively on these datasets, motivating the use of a Gaussian process objective for alignment. Furthermore, we see that our non-parametric approach to modelling the time warps improves the flexibility of the model; out of the two models that rely on energy minimisation as the alignment objective, \emph{energy+basis} and \emph{energy+gplvm}, the latter one demonstrates lower warping error and significantly lower standard deviation on this dataset. This result supports the premise that even though the non-parametric representation allows for any smooth monotonic warp, the probabilistic framework places sufficient structure to make the problem well posed and avoid over-fitting. An example of warps and alignments for this experiment are available in the supplementary material.

\paragraph{Dataset for clustering}
In our second experiment, we consider a dataset that contains multiple clusters of sequences. This task requires the sequences to be aligned within each cluster. None of PDTW, PCTW, GTW nor the energy minimisation methods are able to perform this task as they have no knowledge of the underlying structure of the dataset. The SRVF algorithm performs clustering by first aligning the data in terms of amplitude and phase, then performing fPCA based on the estimated summary statistics, and finally modelling the original data using joint Gaussian or non-parametric models on the fPCA representations. We compare the performance of the SRVF algorithm with our approach as well as the variant of our approach with fixed basis functions.

\begin{figure}[t!]
\centering
        \begin{subfigure}[h]{0.48\textwidth}
                \includegraphics[width=\textwidth, height=7.cm]{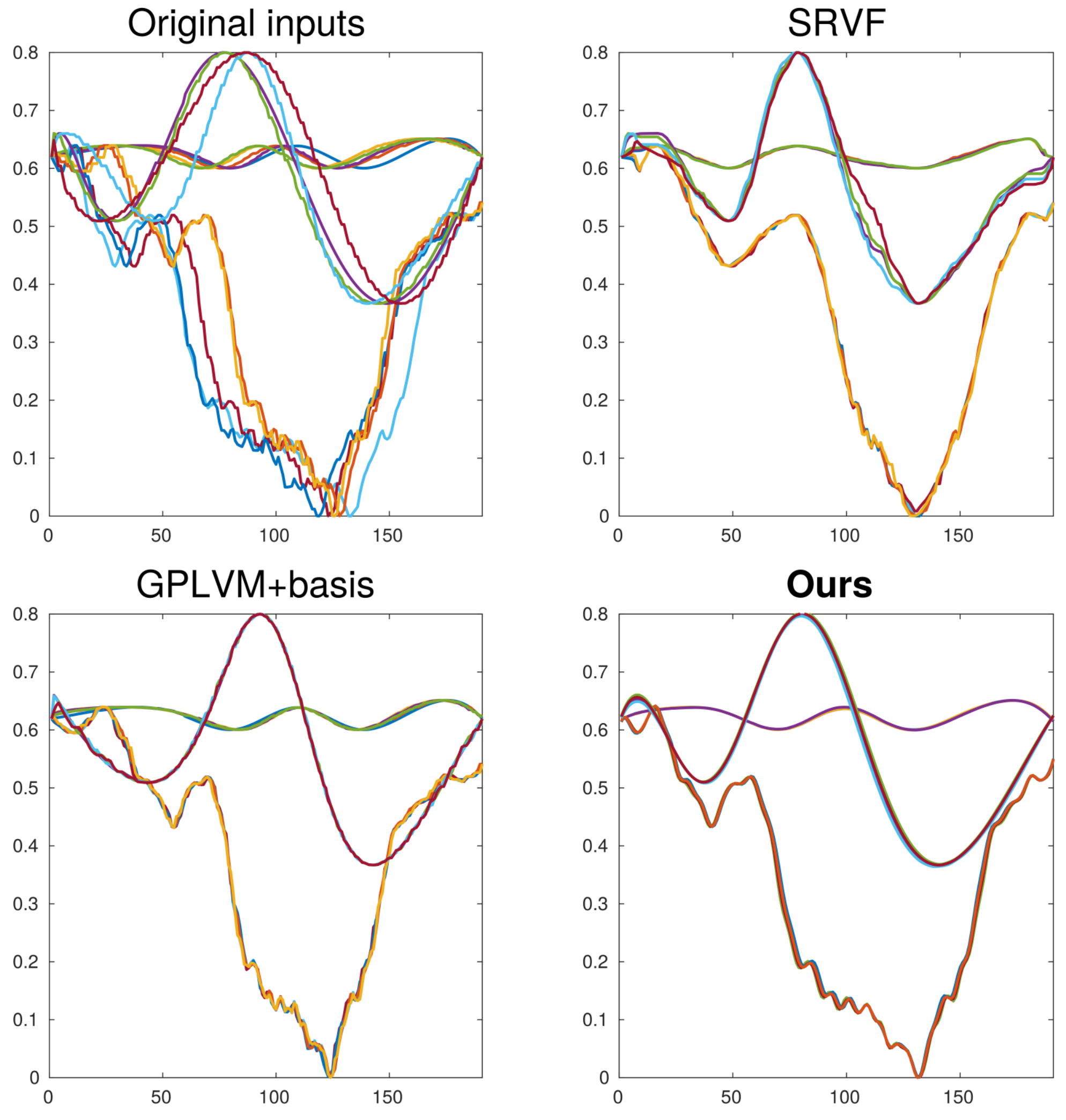}%
        \end{subfigure} % \\ \vspace{10pt}
        \begin{subfigure}[h]{0.48\textwidth}
                \includegraphics[width=\textwidth, height=7.cm]{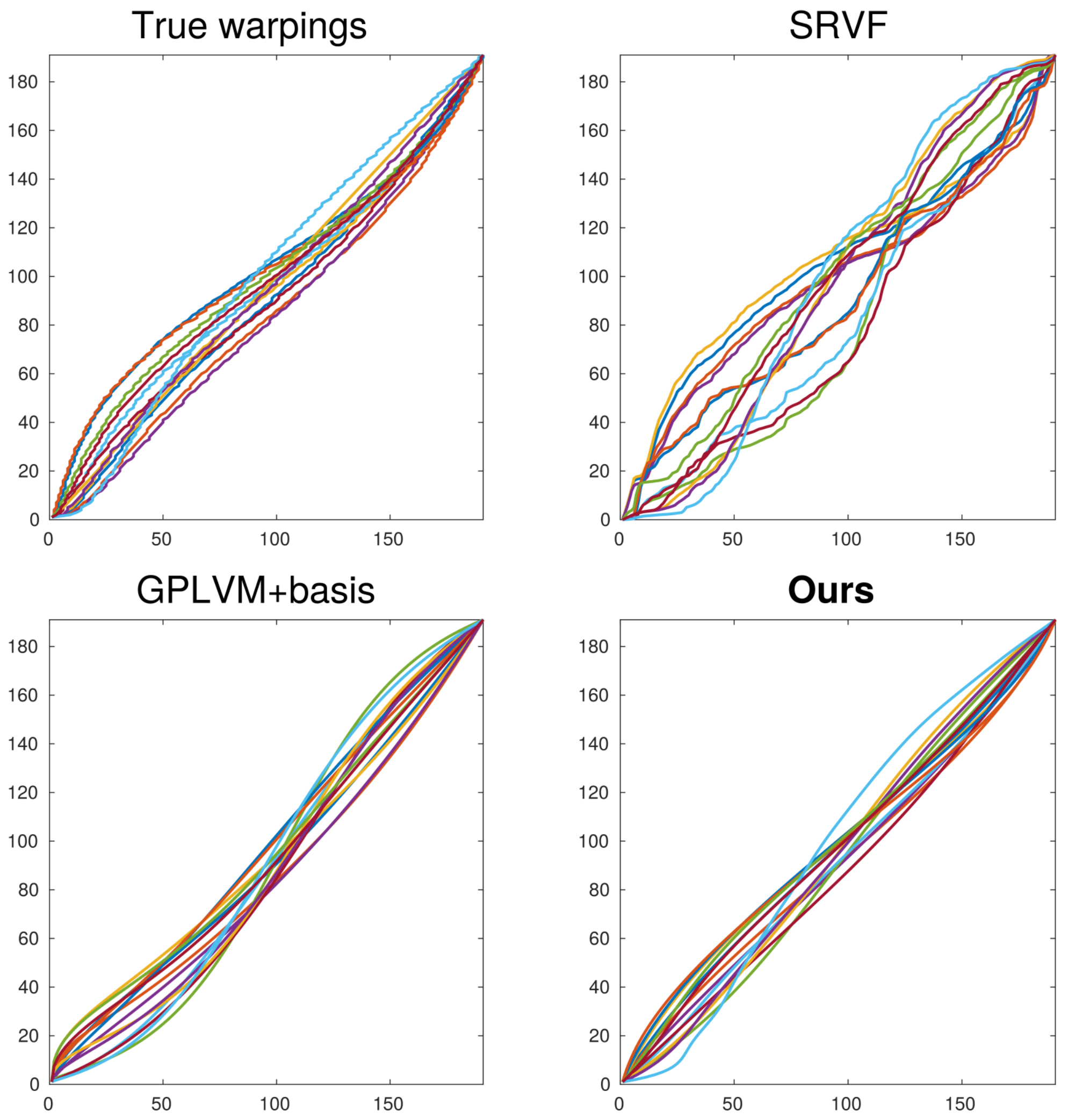}%
        \end{subfigure} 
                \caption{Alignment of $15$ sequences that belong to $3$ different clusters (top 4 graphs) and the corresponding warping functions.} \label{fig:clusters_align}      
\end{figure}

We consider a dataset that contains three distinct groups of functions that were generated by temporally transforming three random 2D curves as described previously. All three approaches rely on the structure of the data alone to recognise the existence of the clusters and Fig.~\ref{fig:clusters_align} shows that all three methods are able to align the data within clusters. 
\begin{figure*}[t!]
\centering
\begin{minipage}{\textwidth}
  \centering
  \begin{subfigure}[h]{0.31\textwidth}
                \includegraphics[width=\textwidth, height=3.3cm]{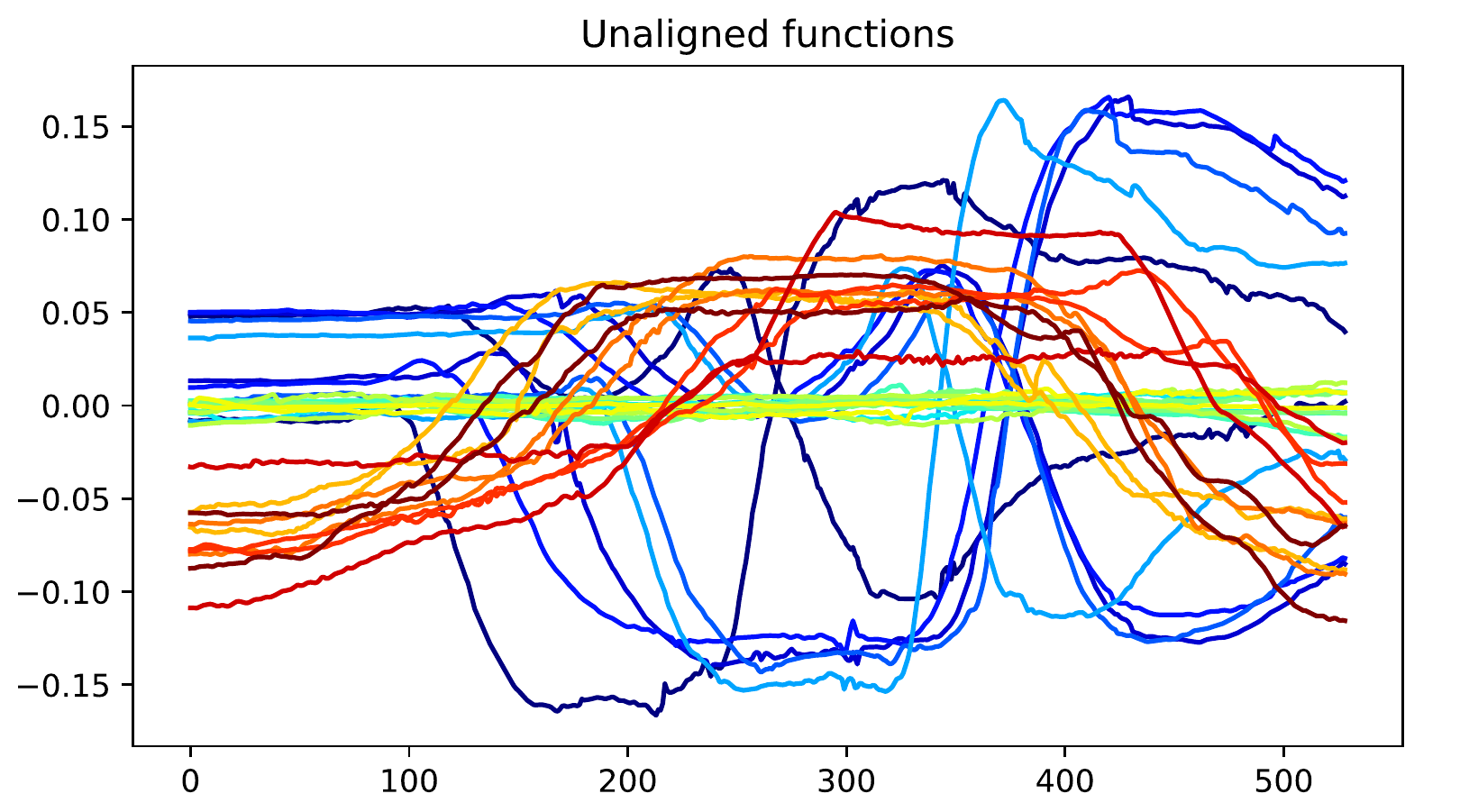}
        \end{subfigure} %\\ \vspace{1pt}
        \begin{subfigure}[h]{0.34\textwidth}
                \includegraphics[width=\textwidth, height=3.3cm]{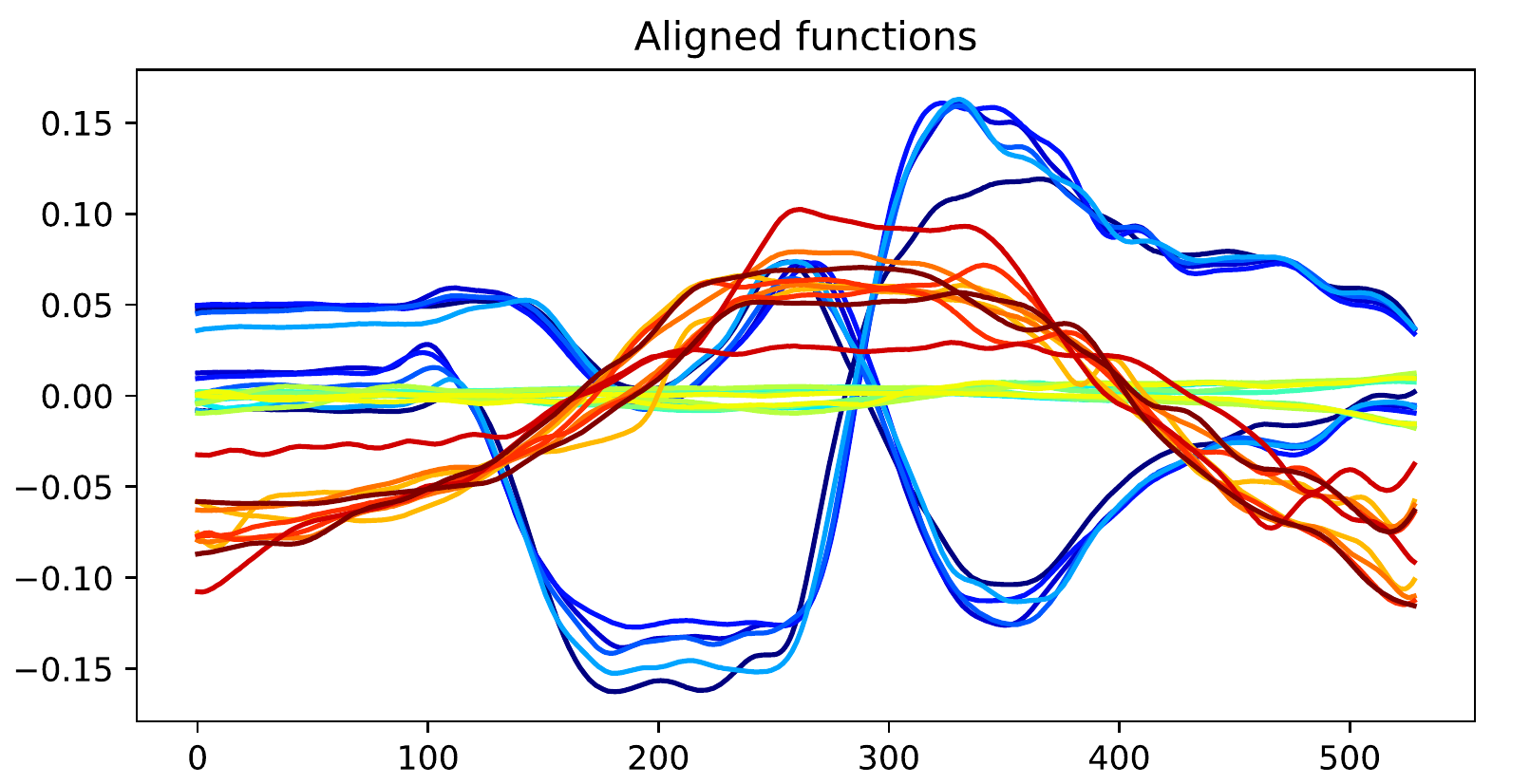}
        \end{subfigure} %\\ \vspace{1pt}
        \begin{subfigure}[h]{0.34\textwidth}
                \includegraphics[width=\textwidth, height=3.3cm]{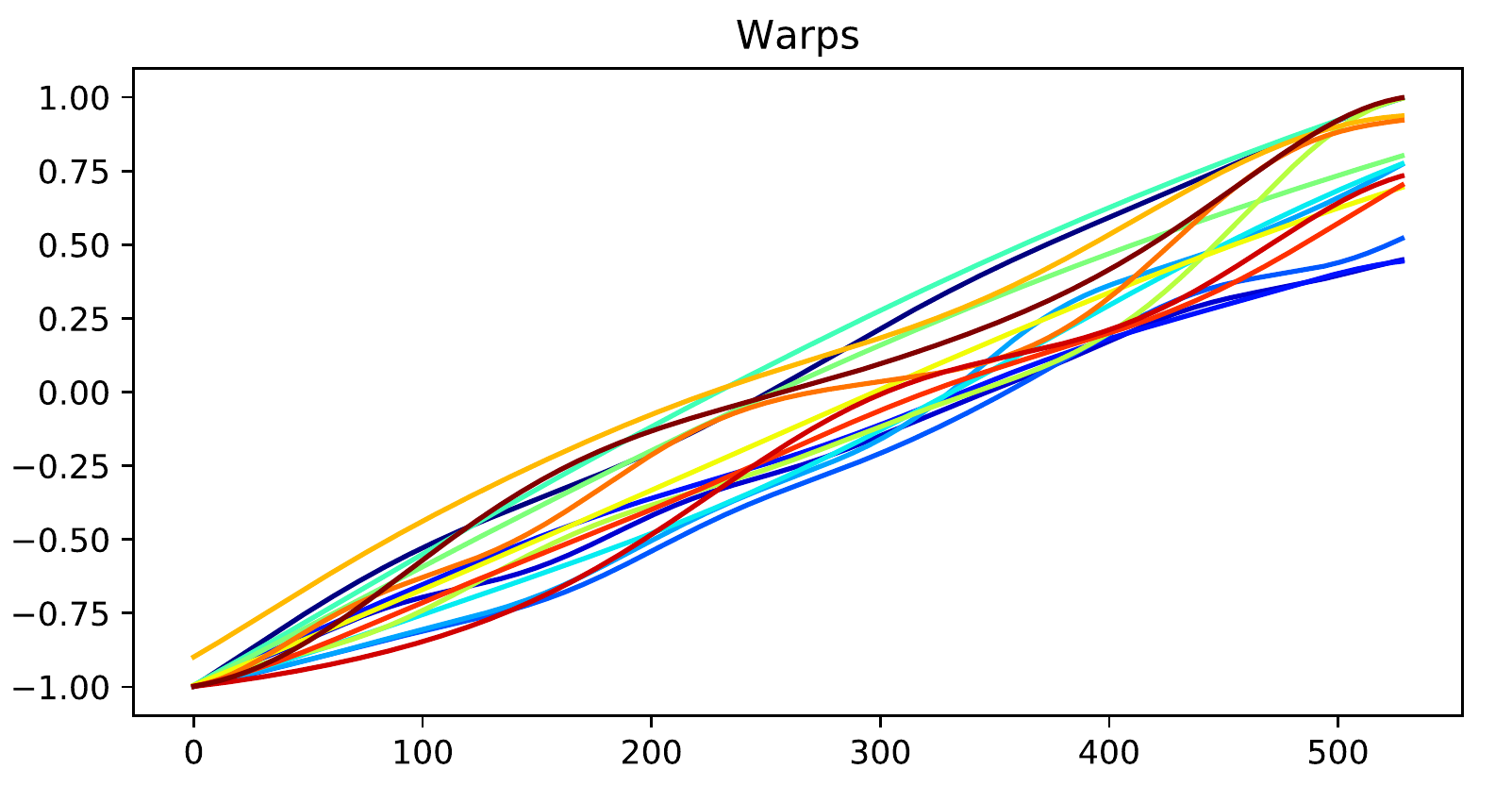}
        \end{subfigure}\\
        \caption{GP-LVM alignment demonstrates the preference for a simplified explanation when the model is given the ability to align the data.} \label{fig:exp2}   
\end{minipage}%
\hspace{5pt}
\begin{minipage}{\textwidth}
  \centering
  \begin{subfigure}{0.49\textwidth}
            \centering
                \includegraphics[height=3.2cm]{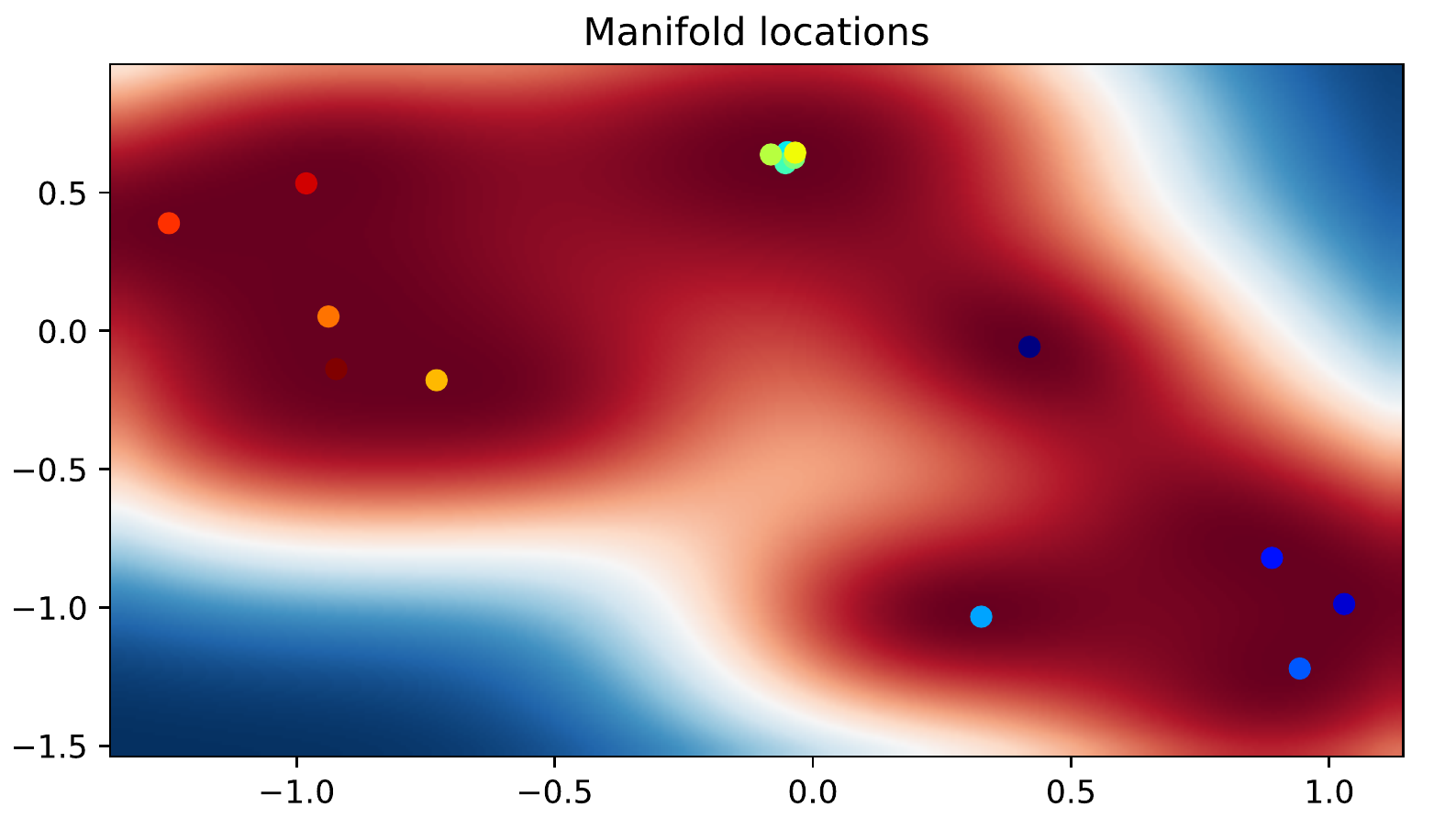}
                \caption{Without alignment.} 
        \end{subfigure} % \\ \vspace{8pt}
        \begin{subfigure}{0.49\textwidth}
        \centering
                \includegraphics[height=3.2cm]{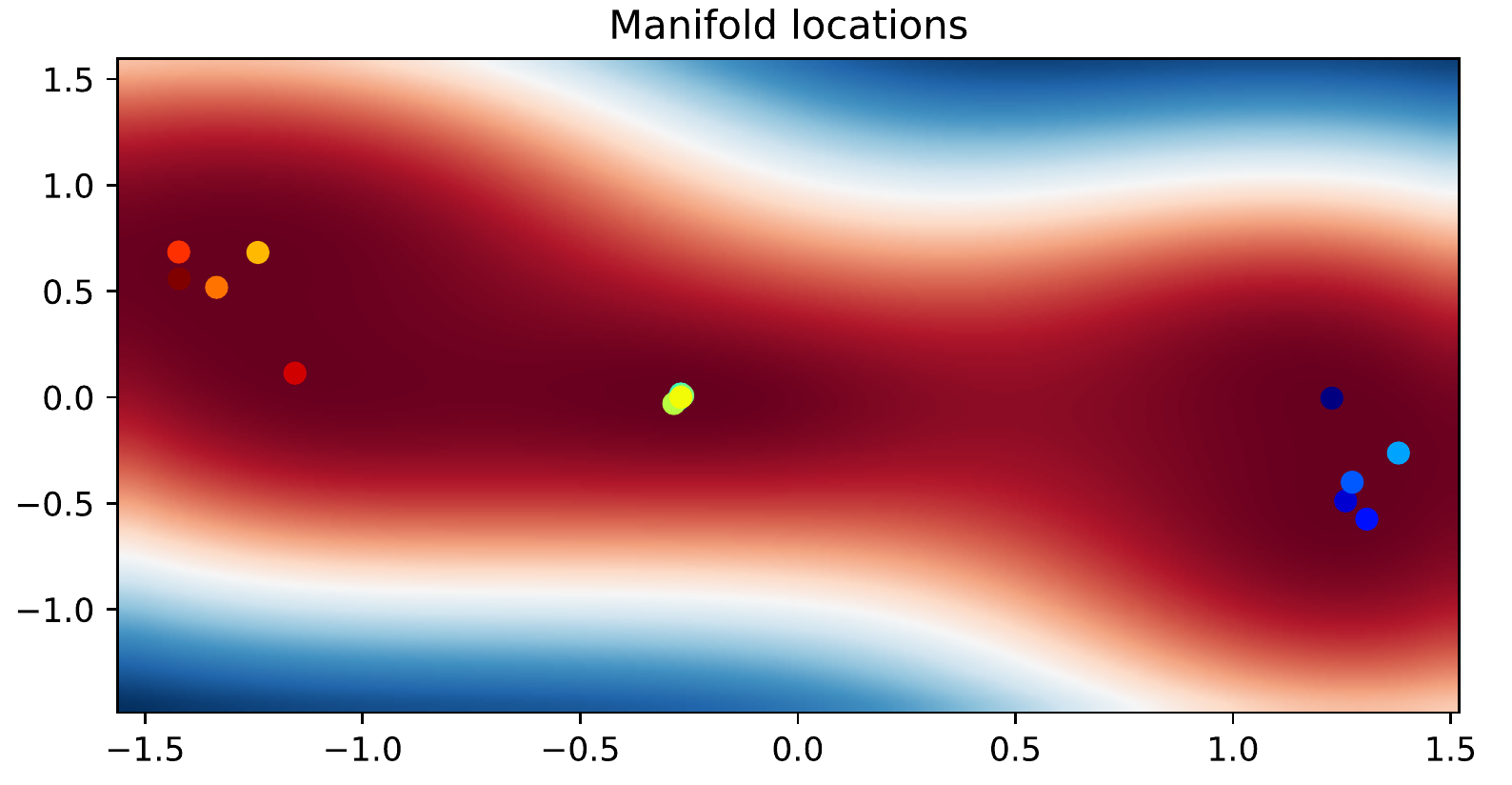}
                \caption{With alignment.} 
\end{subfigure}
\caption{2D manifolds produced without and with alignment in the GP-LVM. Using the alignments emphasizes the existence of multiple clusters of data and aligns data points within each cluster.}
                 \label{fig:manifolds}
\end{minipage}
\end{figure*}
\begin{figure*}[t]
\centering
        \begin{subfigure}[t]{0.39\textwidth}\centering
                \includegraphics[height=2.8cm]{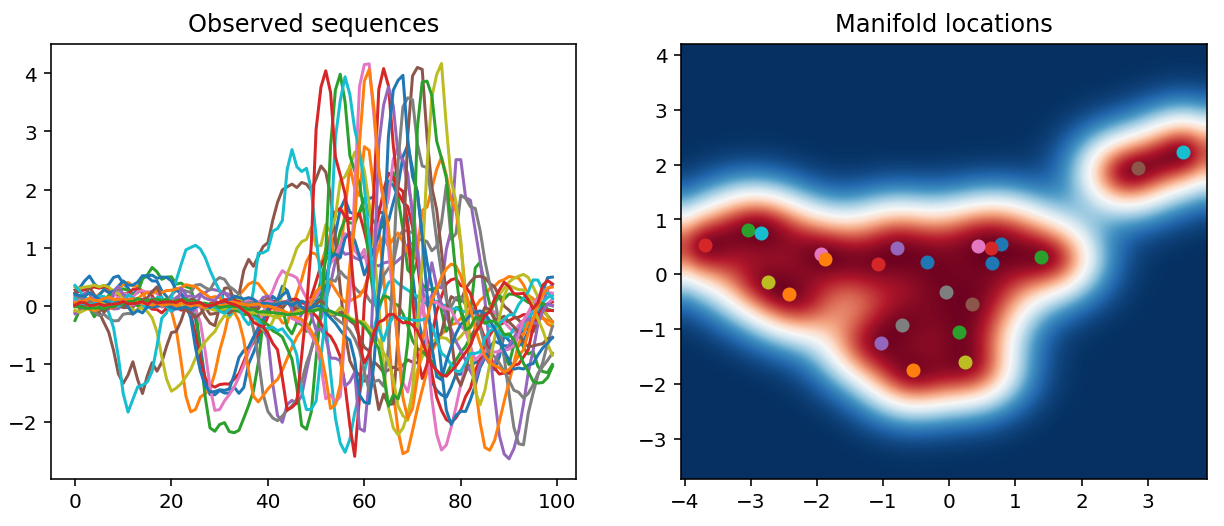} %[height=0.3\textheight]
        \caption{The clustering of the unaligned observed sequences does not reveal the two types of heartbeats.} \end{subfigure}\hfill%
        \begin{subfigure}[t]{0.59\textwidth}\centering
                \includegraphics[height=2.8cm]{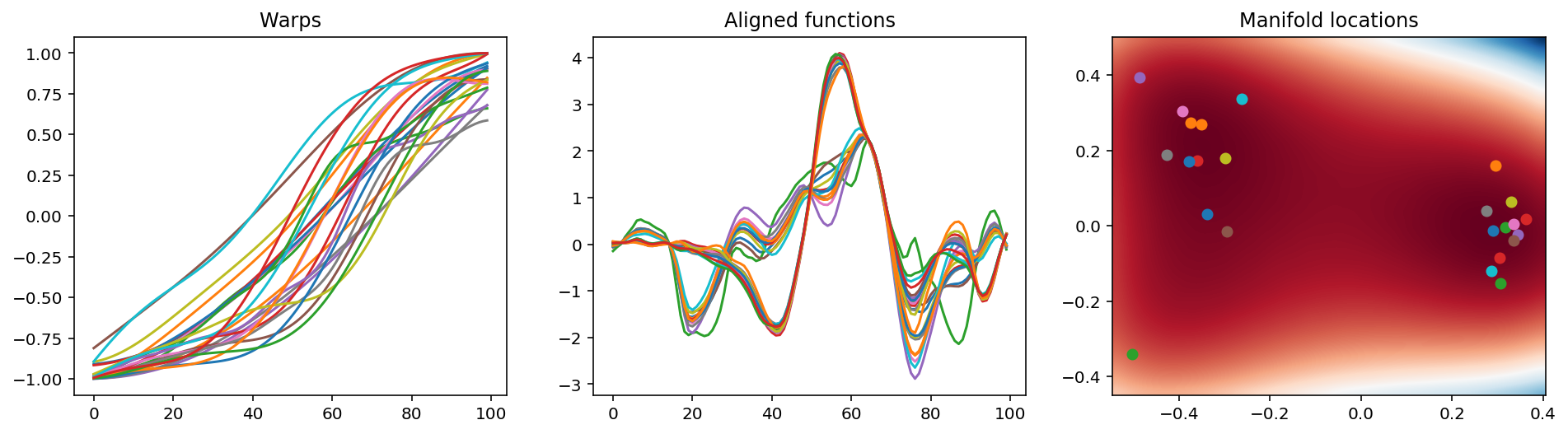} %[height=0.3\textheight]
        \caption{Accounting for the alignment of sequences allows us to discover automatically the two different types of heartbeats.} \end{subfigure}
\caption{Alignment of heartbeats data~\cite{Bentley:2011}.}\label{fig:heartbeats}
\end{figure*}
The performance of the methods is contrasted by calculating the MSE among all pairs of sequences within each group (alignment error) and the MSE between the true warping functions and the warping functions calculated using each of the methods (warping error). For this comparison we repeat the test $25$ times with randomly selected initial curves, number of dimensions and number of sequences per group. The quantitative comparison in Table~\ref{table:quantitative} shows that our method consistently achieves the lowest alignment errors (i.e. with lowest standard deviation (SD) on the set of datasets). 

Our method, as well as the parametric variant of it, also achieves low warping errors in comparison to SRVF which implies that they are able to reconstruct the original temporal transformations more accurately than SRVF. This behaviour is apparent in Fig.~\ref{fig:clusters_align} where the warping functions produced by our method, and the parametric version of it, resemble the true warps while SRVF estimates noticeably different warping functions; this results in unpredictable distortions in the aligned dataset. These results reflects the differences between the SRVF method and our approach; while SRVF is cast as an optimisation problem over a constrained domain, the domain of our probabilistic formulation is much larger but, importantly, structured from the assumptions encoded in the prior. This provides a better regularisation ultimately leading to the improvement in the recovered warpings.

\paragraph{Motion capture data}We evaluate the performance of our model on a set of motion capture data from the CMU database~\cite{CMU:mocap}, where each input sequence corresponds to a short clip of motion and the data is represented as quaternion locations of the joints of the subject performing the motion. We use the motion of subject no $64$ from the CMU dataset that correspond to golf related motions such as a swing, a putt, and placing and picking up of a ball. We consider five instances of three different motions that need to be temporally aligned within the three groups. Fig.~\ref{fig:exp2} illustrates how our model favours the simplified, \ie~aligned, inputs. The corresponding manifolds produced using a traditional GP-LVM (\ie~without alignment) and a manifold produced using our approach are shown in Fig.~\ref{fig:manifolds}. Our model produces a fine alignment of the input sequences within each of the groups, and consequently the resulting two-dimensional manifold offers a good separation of the three groups. We note that the manifold produced using GP-LVM without alignment contains more isolated areas, which means the model is less capable of generalising between the warps. Therefore, our implicitly aligned model is able to generate smoother transitions in the manifold, producing high quality predicted outputs of novel alignments.

\paragraph{Heartbeats data} This dataset contains heartbeat sounds, and it is known that a normal heart sound has a clear "lub dub, lub dub” pattern which varies temporally depending on the age, health, and state of the subject~\cite{Bentley:2011}. Our approach automatically aligns and clusters the heart sounds recorded by a digital stethoscope. Instead of using a pre-processing step with a low-pass filter to account for the noise in the high frequencies, we use a Mat\'{e}rn $3/2$ kernel that takes into count the rapid variations in the recordings while also limiting the effect of the uninformative high frequency noise. Fig.~\ref{fig:heartbeats} illustrates how simultaneous fitting and alignment allows us to correctly discover and cluster the two types of heartbeats.
% -----------------------------------------------------------------------------------------

%% file: includes/conclusions.tex
% \flushcolsend
\section{Conclusion and Future Work} 
\label{sec:conclusions}
We have presented a probabilistic model that is able to implicitly align inputs that contain temporal variations. Our approach models the observed data directly producing a generative model of the functions rather than interpolating between observations. In addition, using a GP-LVM for alignment builds an unsupervised generative model that has the benefit of simultaneous clustering and aligning the input sequences. Furthermore, we proposed a continuous, non-parametric explicit model of the time warping functions that removes issues such as quantisation artefacts and the need for ad-hoc pre-processing. We demonstrated that the proposed approaches perform competitively on alignment tasks, and outperform the existing methods on the task of simultaneous alignment and clustering. In the future we will consider the use of Bayesian GP-LVM for automatic model selection and will test the framework on additional datasets, including multi-modal data.

%% file: includes/appendix_old.tex
% !TEX root = ../aistats_supp.tex
%\subsection*{Datasets with quantifiable comparisons}
\subsection*{Motion capture dataset}
\begin{figure}[h]
\centering
\includegraphics[width=0.49\textwidth]{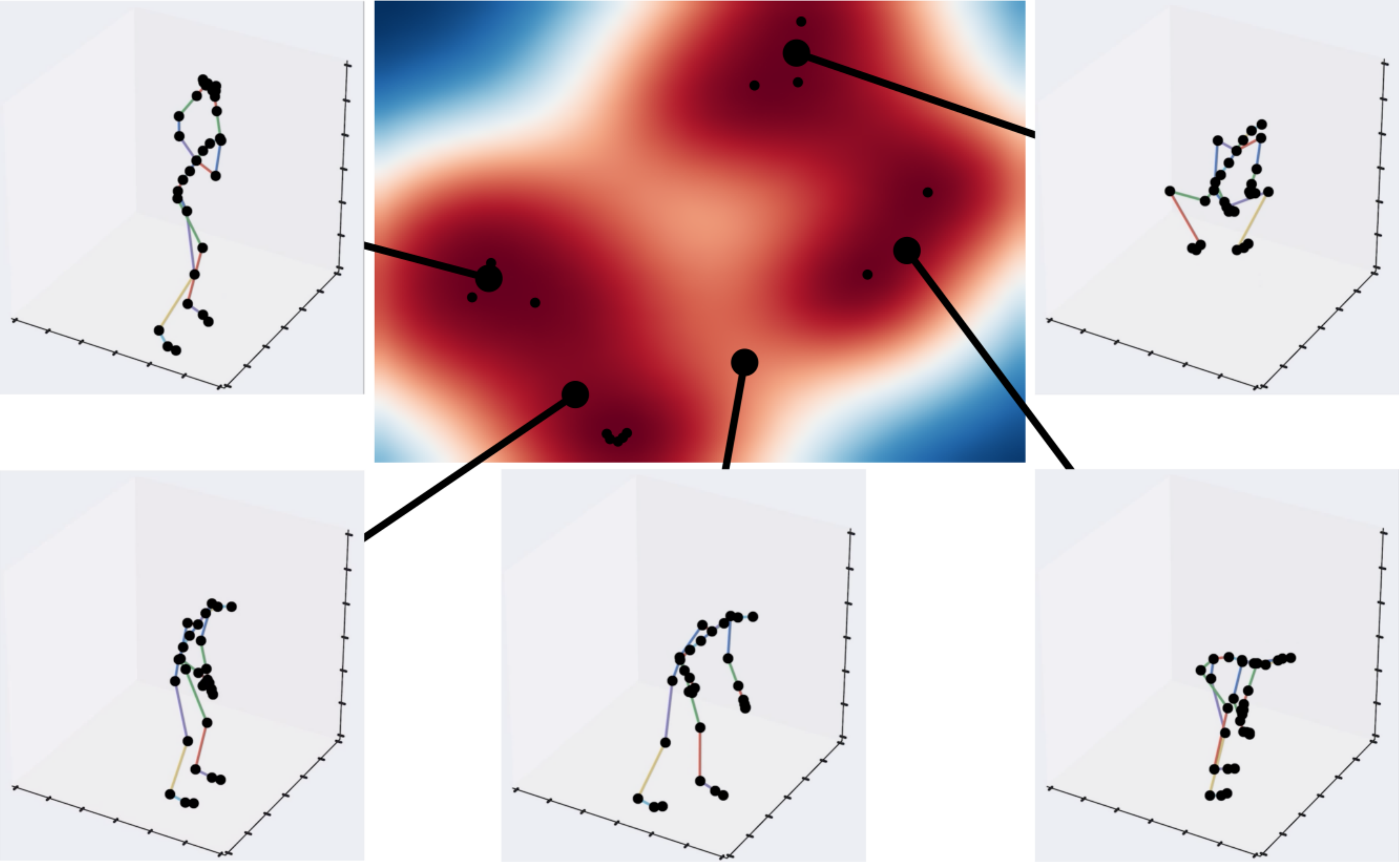}
                 \caption{An advantage of our approach is that it not only aligns the data but is also a generative probabilistic model. Here we show novel sequences generated at new locations in the manifold. The black dots indicate the embedded locations of the training sequences. We note that, while we have only shown still images, each manifold location describes an entire time series. A video showing this is included with the supplementary material.} \label{fig:gen_model} % 
\end{figure}

In this experiment we use the full set of joint motions to align a set of sports actions (see \S 4 for further information on the motion capture dataset). In Fig.~\ref{fig:gen_model} we provide an illustration of the power of using a generative model for alignment. New locations in the manifold encode novel motion sequences that are supported by the data. By allowing the model to align the data, it greatly improves the generative power as the model is capable of producing a wider range of plausible motions. 

\begin{figure*}[h]
\begin{minipage}{0.49\textwidth}
\centering
        \begin{subfigure}[h]{\textwidth}
                \includegraphics[width=\textwidth]{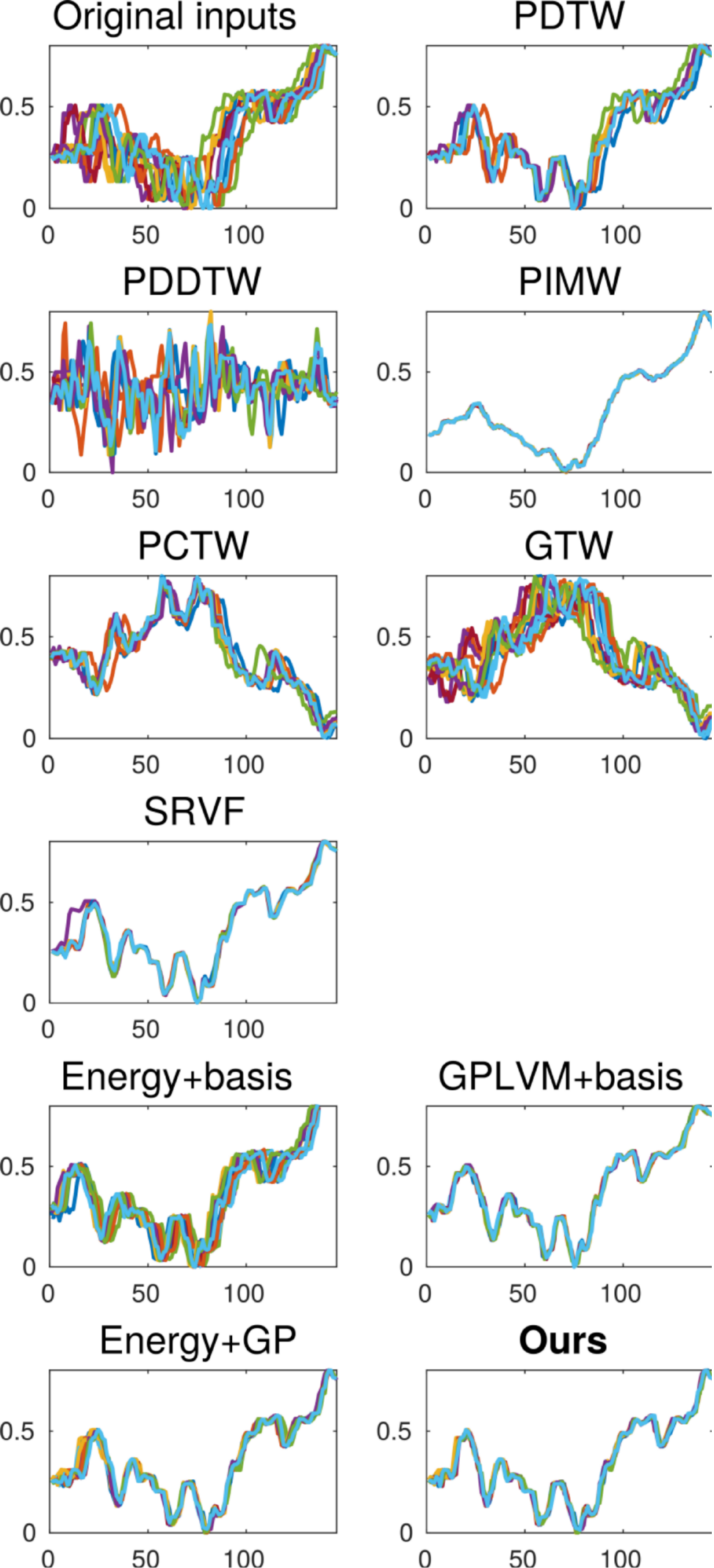}
        \end{subfigure}
                \caption{Original inputs and aligned sequences estimated by DTW, DDTW, IMW, CTW, GTW, SRVF, our approach and its three variants.} \label{fig:toys_aligns}        
\end{minipage}%
\hspace{4pt}
\begin{minipage}{0.49\textwidth}
\centering
        \begin{subfigure}[h]{\textwidth}
                \includegraphics[width=\textwidth]{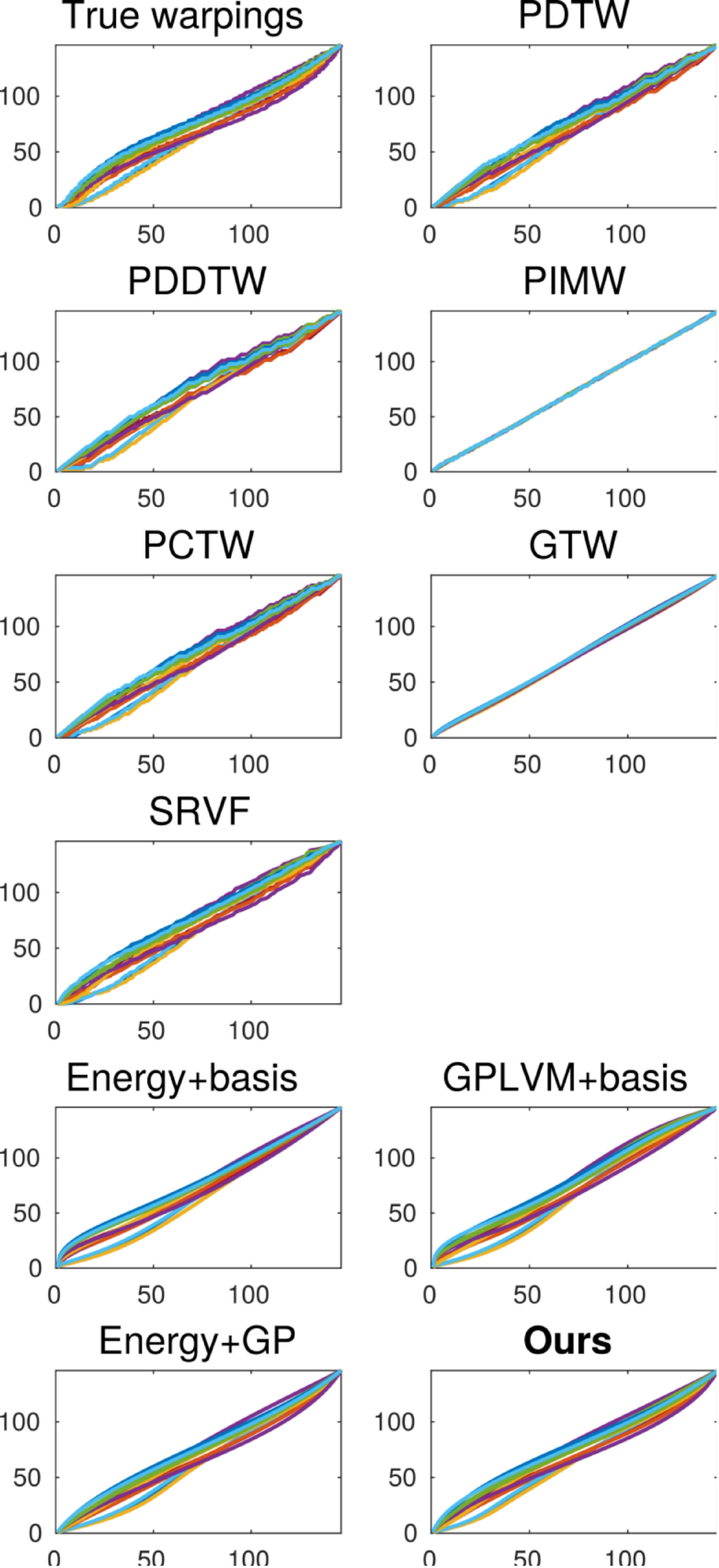}
        \end{subfigure}
                \caption{True warps and warps estimated by DTW, DDTW, IMW, CTW, GTW, SRVF, our approach and its three variants.} \label{fig:toys_warps}        
\end{minipage}
\end{figure*}

Fig.~\ref{fig:toys_aligns} and Fig.~\ref{fig:toys_warps} give an example of the alignments and the warps produced by our method on the quantifiable dataset, see \S 4 in the paper. The detailed results of our experiments on this dataset are provided in Table~\ref{table:main_results}.

\subsection*{iPhone motion data}
This dataset contains aerobic actions recorded using the Inertial Measurement Unit (IMU) on a smartphone~\cite{McCall:2012}, which contain high frequency variations. Unlike previous methods~\cite{Tucker:2013}, which require the data to be smoothed first, our framework allows us to take into account the prior belief about the dataset in a principled way. By replacing the smooth RBF kernels for modeling the data with a Mat\'{e}rn $1/2$ kernel and taking into account the periodic nature of the actions by also including an additive periodic kernel, we are able to model the data without the need for preprocessing. Furthermore, by removing the smoothing prior from the warping functions, we allow the warps to be more flexible improving the alignment accuracy. 

\begin{figure*}[h]
\centering
        \begin{subfigure}[h]{\textwidth}
                    \includegraphics[width=\textwidth]{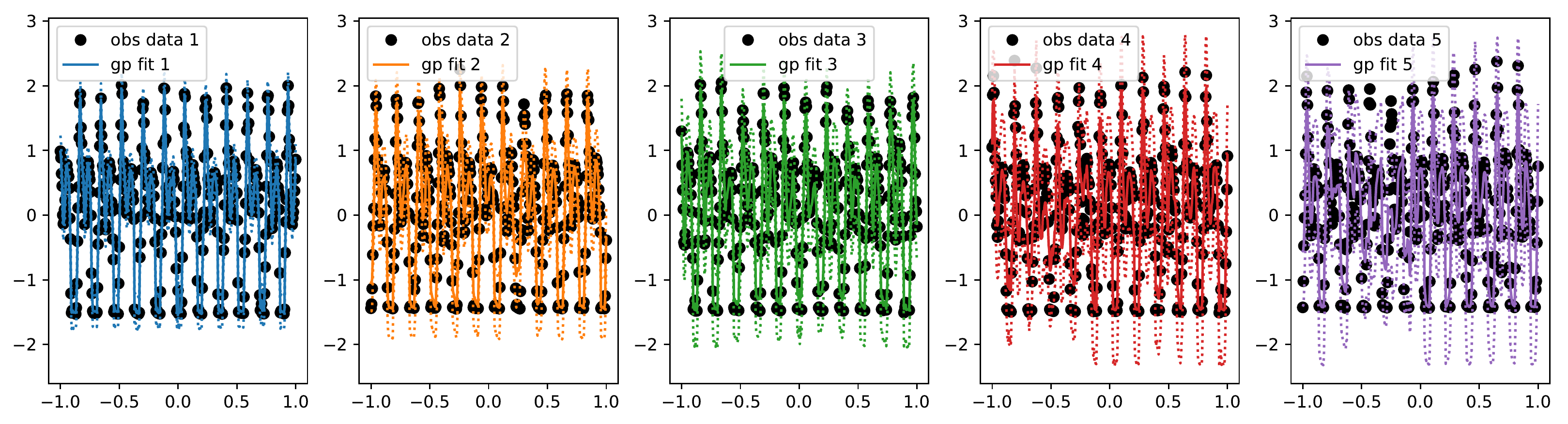} %height=3.5cm
        \end{subfigure}  \\ \vspace{5pt}
        \begin{subfigure}[h]{\textwidth}
                \includegraphics[width=\textwidth]{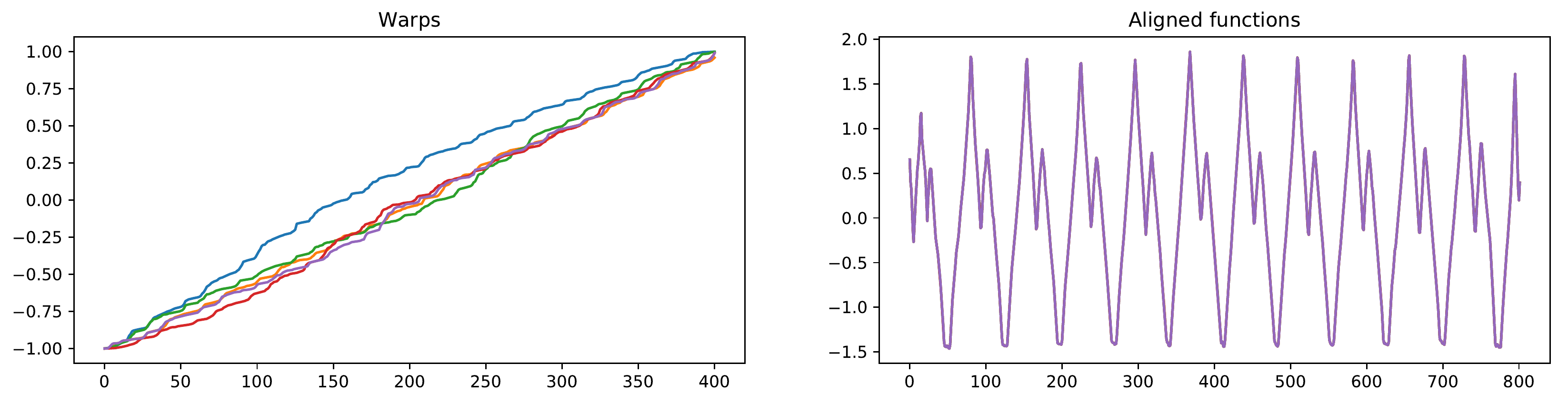} %height=3.5cm
        \end{subfigure}
        \caption{The top row shows the observed data and the fitted Gaussian Processes. The bottom row shows the corresponding warps (left) and the aligned functions.}\label{fig:gait}
\end{figure*}

The alignment results for the iPhone motion data are shown in Fig.~\ref{fig:gait}. The IMU includes a 3D accelerometer, a gyroscope, and a magnetometer, and records samples at 60 Hz. As in~\cite{Tucker:2013}, for our experiment we take the accelerometer data in the x-direction for the jumping actions from subject $3$, and, in particular, we look at $5$ sequences each of which contains $400$ frames. A Mat\'{e}rn $1/2$ kernel and a periodic kernel are used to fit the sequences as they contain high frequency variations, and we remove the smoothness constraint from the model of the warping functions to allow them to be more flexible.

\subsection*{Shift task}
\begin{figure*}
\centering
        \begin{subfigure}[h]{0.49\textwidth}
                    \includegraphics[height=0.2\textheight]{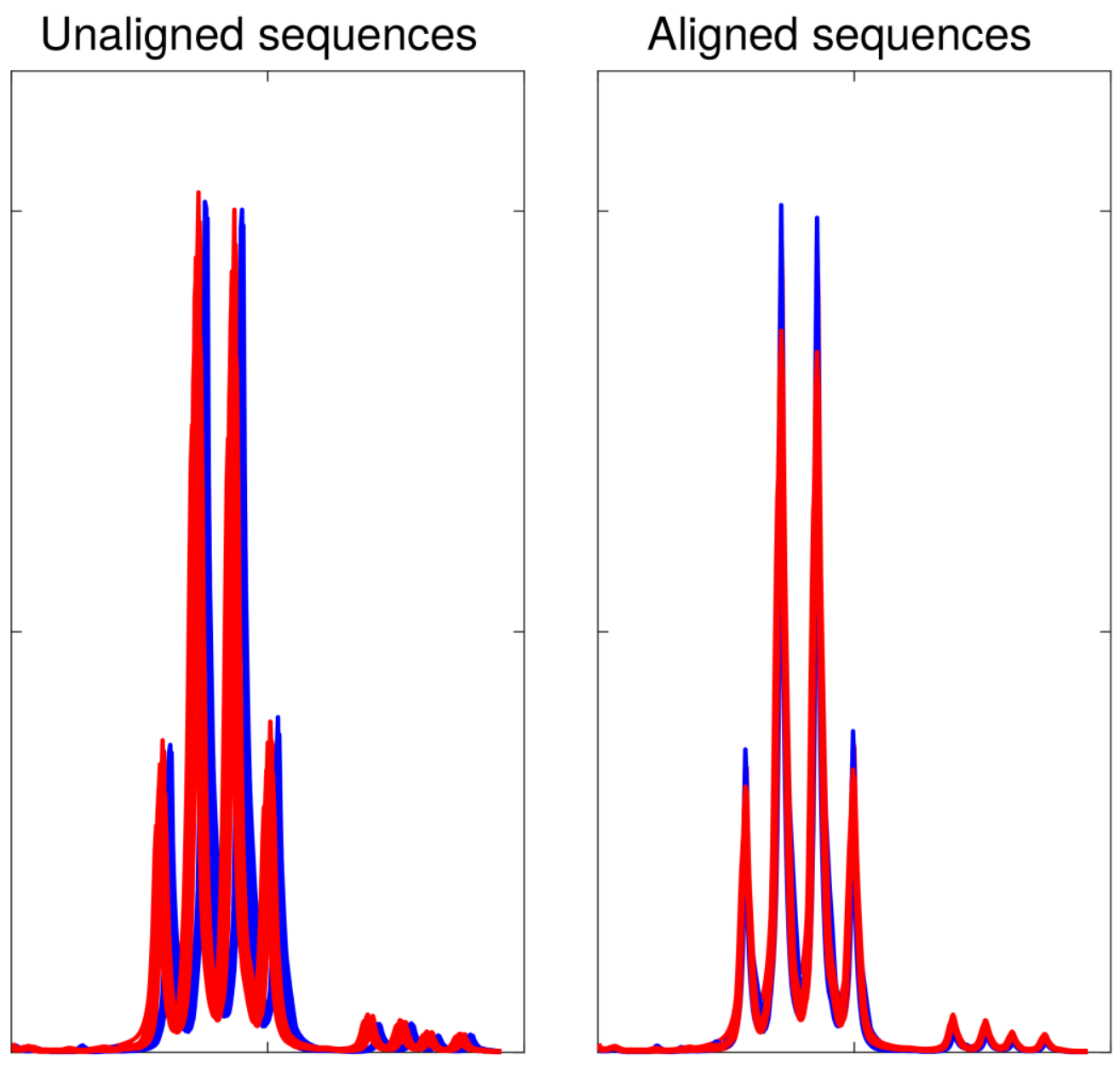}
        \end{subfigure}
        \begin{subfigure}[h]{0.49\textwidth}
                \includegraphics[height=0.2\textheight]{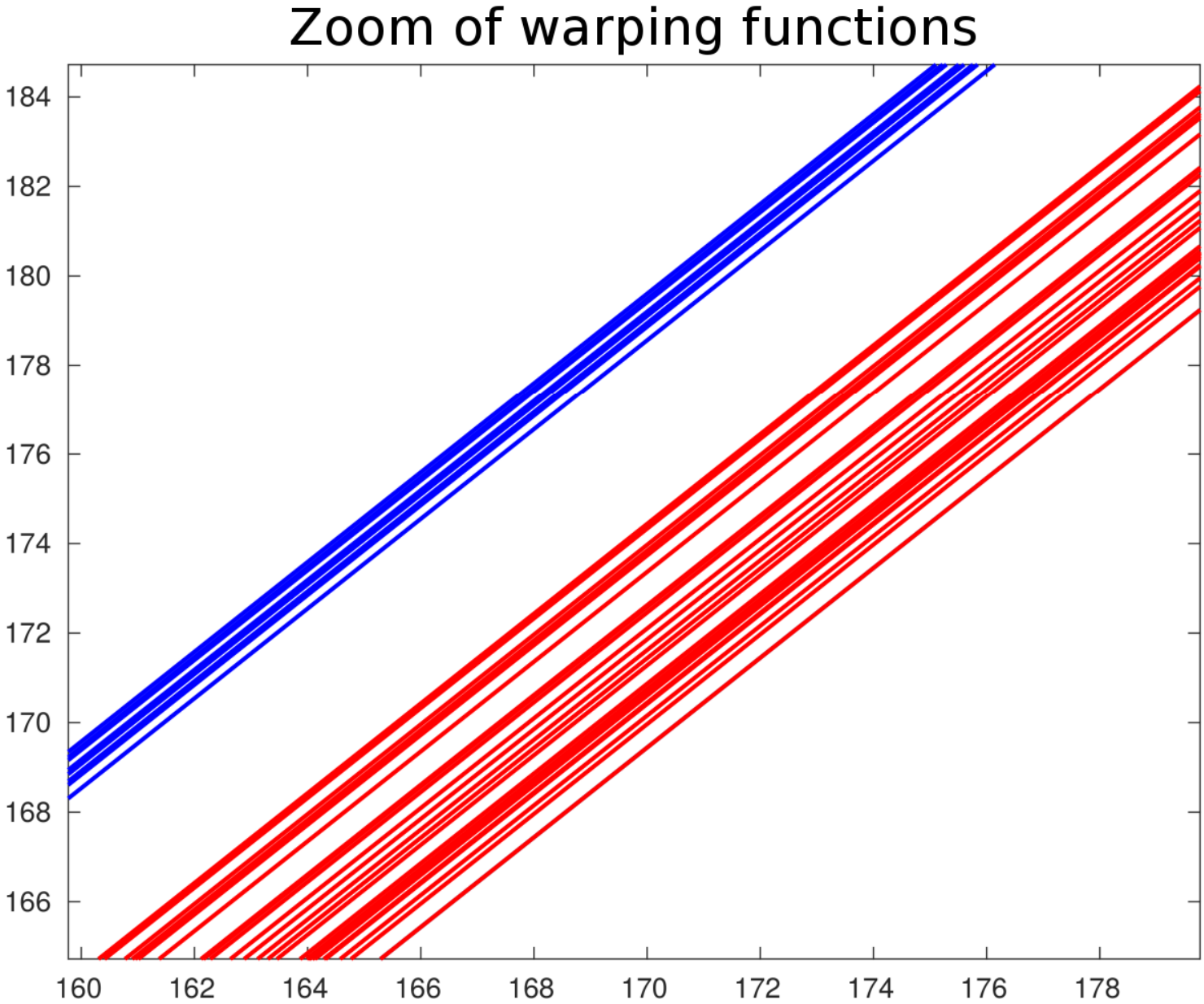}
        \end{subfigure}
	\caption{Alignment of NMR spectrum data~\cite{Marron:2015}. The zoom of the warping functions show the separation of the white/ros\'{e} wines (shown in blue) and red wines (shown in red).}\label{fig:wines}
\end{figure*}

A common task in functional data alignment is that of estimating uniform translations of the time axis. One particular problem described by Marron \emph{et al.} is that of aligning nuclear magnetic resonance (NMR) spectrum corresponding to different chemical components (e.g. ethanol) for a set of wines~\cite{Marron:2015}. It is known that pH differences in wines induce a shift in values of the components and impedes their identification~\cite{Larsen:2006}. As shown by Marron \emph{et al.} the alignment may be achieved using uniform shifts and minimizing the loss that requires sequences to be proportional to each other. Such operation is included in our model allowing us to perform the task of NMR spectrum alignment, and we are able to demonstrate a separation in the phase between the red wines and the white and ros\'{e} wines, see Fig.~\ref{fig:wines}. 

\begin{table*}[h]
\centering
\input{includes/results_table.tex}%
\caption{Datasets used for our evaluation where $J$ and $T$ refer to the number of sequences and dimensionality. Results are presented as MSE of warpings. The summary of the results presented in this table is given in Fig. $3$ in the paper. }
\label{table:main_results}
\end{table*}

%% file: includes/results_table.tex
\scalebox{0.72}{%
{%
\centering
\small%
\renewcommand{\tabcolsep}{2pt}%
\def\arraystretch{1.5}%
\begin{tabular}{c|ccccccccccccccccccccccccc|c}
Dataset no &1          &2          &3          &4          &5          &6          &7          &8          &9          &10         &11         &12         &13         &14         &15         &16         &17         &18         &19         &20         &21         &22         &23         &24         &25         &Mean              \\ \hline
 $J$  & 13 & 10 & 10 & 7  & 13 & 6  & 6  & 12 & 3  & 10 & 13 & 8  & 6  & 10 & 7  & 5  & 5  & 14 & 8  & 3  & 8  & 11 & 7  & 6  & 9  &        \\
 $T$   & 258 & 157 & 107 & 246 & 169 & 131 & 92  & 144 & 138 & 298 & 146 & 240 & 204 & 213 & 157 & 230 & 247 & 196 & 248 & 277 & 141 & 153 & 83  & 285 & 178 &          \\ \hline
 PDTW  & 9.32  & 14.38 & 5.88  & 12.42 & 10.57 & 10.03 & 2.30  & 6.60  & 2.41  & 16.99 & 9.84  & 13.47 & 4.95  & 16.72 & 4.00  & 6.97  & 14.03 & 6.57  & 15.01 & 1.98  & 4.44  & 5.16  & 5.05  & 13.19 & 10.85 & 8.93  \\
 PDDTW & 10.55 & 14.67 & 6.80  & 13.86 & 12.18 & 10.30 & 2.58  & 7.35  & 3.95  & 24.59 & 10.68 & 15.45 & 6.57  & 18.60 & 5.54  & 8.35  & 14.91 & 7.70  & 16.38 & 5.37  & 6.11  & 7.43  & 5.44  & 14.02 & 11.17 & 10.42 \\
 PIMW  & 26.61 & 19.27 & 13.36 & 24.59 & 21.16 & 14.75 & 6.48  & 22.14 & 5.36  & 38.54 & 18.06 & 26.72 & 16.77 & 28.39 & 9.54  & 22.31 & 21.23 & 22.35 & 27.54 & 11.13 & 12.05 & 18.45 & 9.06  & 30.40 & 16.54 & 19.31 \\
 PCTW  & 12.12 & 15.30 & 9.50  & 18.89 & 15.45 & 11.32 & 2.77  & 10.58 & 2.66  & 17.26 & 9.98  & 24.72 & 8.04  & 17.19 & 6.23  & 10.21 & 16.03 & 10.73 & 15.17 & 7.12  & 5.88  & 5.59  & 6.00  & 16.92 & 10.98 & 11.47 \\
 GTW   & 6.52  & 9.54  & 6.54  & 6.96  & 7.50  & 3.23  & 4.27  & 10.63 & 0.86  & 33.31 & 2.55  & 21.28 & 5.35  & 4.72  & 3.38  & 23.28 & 8.50  & 10.70 & 2.83  & 3.93  & 2.74  & 2.56  & 5.45  & 15.70 & 2.63  & 8.20  \\
 SRVF & 6.74 & 3.41 & 4.73 & 8.06 & 6.94 & 4.14 & 2.14 & 3.34 & 3.10 & 8.65 & 5.13 & 5.33 & 3.57 & 7.37 & 5.78 & 7.02 & 4.87 & 4.71 & 5.54 & 5.12 & 3.82 & 4.55 & 2.22 & 4.53 & 7.04 & 5.11 \\ \hline
 energy+basis & 9.91         & 5.08         & 4.69         & 6.12         & 6.89         & 3.06         & 2.10         & 5.33         & 0.98         & 14.45        & 6.64         & 6.57         & 2.10         & 10.38        & 2.74         & 3.83         & 5.13         & 7.94         & 5.67         & 1.37         & 4.42         & 5.79         & 2.12         & 5.00         & 4.88         & 5.33         \\
 gplvm+basis   & 5.85          & 3.09          & 2.98          & 5.29          & 3.65          & 2.24          & 1.31          & 3.53          & 0.87          & \textbf{1.85} & 3.67          & \textbf{3.68} & \textbf{1.49} & 5.58          & 1.58          & 3.55          & 3.69          & \textbf{2.13} & 3.12          & \textbf{1.22} & 3.20          & 3.59          & 1.35          & \textbf{1.72} & \textbf{1.90} & 2.88          \\
 energy+gplvm  & 6.80          & \textbf{2.45} & 3.29          & 6.35          & 5.31          & 2.58          & 1.39          & \textbf{3.23} & 0.97          & 4.54          & 2.97          & 4.34          & 2.79          & 3.37          & 3.16          & 3.22          & 4.12          & 3.48          & 2.64          & 2.07          & 3.94          & 2.69          & 2.20          & 2.18          & 2.65          & 3.31          \\ \hline
 \textbf{ours}  & \textbf{4.39} & 4.55          & \textbf{1.79} & \textbf{1.93} & \textbf{2.34} & \textbf{1.91} & \textbf{1.23} & 4.68          & \textbf{0.84} & 3.49          & \textbf{2.11} & 4.94          & 3.47          & \textbf{3.14} & \textbf{1.40} & \textbf{3.03} & \textbf{2.01} & 2.57          & \textbf{2.10} & 1.48          & \textbf{2.20} & \textbf{1.93} & \textbf{1.31} & 2.87          & 2.11           & \textbf{2.55}         \\[-5pt]            
\end{tabular}%
}%
}%

%% file: alignment_aistats.bbl
\begin{thebibliography}{10}

\bibitem{Tensorflow}
Mart\'{\i}n Abadi, Ashish Agarwal, Paul Barham, Eugene Brevdo, Zhifeng Chen,
  Craig Citro, Greg~S. Corrado, Andy Davis, Jeffrey Dean, Matthieu Devin,
  Sanjay Ghemawat, Ian Goodfellow, Andrew Harp, Geoffrey Irving, Michael Isard,
  Yangqing Jia, Rafal Jozefowicz, Lukasz Kaiser, Manjunath Kudlur, Josh
  Levenberg, Dan Man\'{e}, Rajat Monga, Sherry Moore, Derek Murray, Chris Olah,
  Mike Schuster, Jonathon Shlens, Benoit Steiner, Ilya Sutskever, Kunal Talwar,
  Paul Tucker, Vincent Vanhoucke, Vijay Vasudevan, Fernanda Vi\'{e}gas, Oriol
  Vinyals, Pete Warden, Martin Wattenberg, Martin Wicke, Yuan Yu, and Xiaoqiang
  Zheng.
\newblock {TensorFlow: Large-Scale Machine Learning on Heterogeneous Systems},
  2015.
\newblock Software available from tensorflow.org.

\bibitem{Anirudh:2015}
R.~Anirudh, P.~Turaga, J.~Su, and A.~Srivastava.
\newblock {Elastic Functional Coding of Human Actions: From Vector-Fields to
  Latent Variables}.
\newblock In {\em International Conference on Computer Vision and Pattern
  Recognition (CVPR)}, 2015.

\bibitem{Baisero:2015tm}
A.~Baisero, F.~T. Pokorny, and C.~H. Ek.
\newblock {On a Family of Decomposable Kernels on Sequences}.
\newblock {\em arXiv.org}, abs/1501.06284, 2015.

\bibitem{Bentley:2011}
P.~Bentley, G.~Nordehn, M.~Coimbra, and S.~Mannor.
\newblock The {PASCAL} {C}lassifying {H}eart {S}ounds {C}hallenge 2011
  {(CHSC2011)} {R}esults.
\newblock http://www.peterjbentley.com/heartchallenge.

\bibitem{Berndt:1994}
D.~J. Berndt and J.~Clifford.
\newblock {Using Dynamic Time Warping to Find Patterns in Time Series}.
\newblock In {\em International Conference on Knowledge Discovery and Data
  Mining (KDD)}, 1994.

\bibitem{Campbell:2014}
N.~D.~F. Campbell and J.~Kautz.
\newblock {Learning a Manifold of Fonts}.
\newblock {\em ACM Transactions on Graphics}, 33(4), 2014.

\bibitem{Cui:2014}
Z.~Cui, H.~Chang, S.~Shan, and X.~Chen.
\newblock {Generalized Unsupervised Manifold Alignment}.
\newblock In {\em Advances in Neural Information Processing Systems (NIPS)},
  2014.

\bibitem{Cuturi:2011}
M.~Cuturi.
\newblock {Fast Global Alignment Kernels}.
\newblock In {\em International Conference on Machine Learning (ICML)}, 2011.

\bibitem{Cuturi:2007}
M.~Cuturi, J.~P. Vert, O.~Birkenes, and T.~Matsui.
\newblock {A Kernel for Time Series Based on Global Alignments}.
\newblock In {\em International Conference on Acoustics, Speech and Signal
  Processing (ICASSP)}, 2007.

\bibitem{Drydmard:2016}
I.~L. Dryden and K.~V. Mardia.
\newblock {\em Statistical Shape Analysis, with Applications in {R}. Second
  Edition.}
\newblock 2016.

\bibitem{Duncker:2018}
Lea Duncker and Maneesh Sahani.
\newblock Temporal alignment and latent gaussian process factor inference in
  population spike trains.
\newblock {\em bioRxiv}, 2018.

\bibitem{Garreau:2014}
D.~Garreau, R.~Lajugie, S.~Arlot, and F.~Bach.
\newblock {Metric Learning for Temporal Sequence Alignment}.
\newblock In {\em {Advances in Neural Information Processing Systems (NIPS)}},
  2014.

\bibitem{Grochow:2004}
K.~Grochow, S.~L. Martin, A.~Hertzmann, and Z.~Popovi\'{c}.
\newblock {Style-based Inverse Kinematics}.
\newblock In {\em ACM SIGGRAPH}, 2004.

\bibitem{Hsu:2005}
Eugene Hsu, Kari Pulli, and Jovan Popovi\'{c}.
\newblock Style translation for human motion.
\newblock {\em ACM Trans. Graph.}, 24(3):1082--1089, 2005.

\bibitem{Haxby:2011}
V.~H. James, J.~S. Guntupalli, A.~C. Connolly, Y.~O. Halchenko, B.~R. Conroy,
  M.~I. Gobbini, M.~Hanke, and P.~J. Ramadge.
\newblock {A Common, High-Dimensional Model of the Representational Space in
  Human Ventral Temporal Cortex}.
\newblock {\em Neuron}, 72(2):404--416, 2011.

\bibitem{Keogh:2001}
E.~J. Keogh and M.~J. Pazzani.
\newblock {Derivative Dynamic Time Warping}.
\newblock In {\em SIAM International Conference on Data Mining}, 2001.

\bibitem{AdamOpt}
D.~P. Kingma and J.~Ba.
\newblock {Adam: A Method for Stochastic Optimization}.
\newblock In {\em International Conference on Learning Representations
  ({ICLR})}, 2014.

\bibitem{Kurtek:2012}
S.~Kurtek, A.~Srivastava, E.~Klassen, and Z.~Ding.
\newblock {Statistical Modeling of Curves using Shapes and Related Features}.
\newblock {\em Journal of the American Statistical Association},
  107(499):1152--1165, 2012.

\bibitem{Kurtek:2011}
S.~Kurtek, A.~Srivastava, and W.~Wu.
\newblock {Signal Estimation Under Random Time-warpings and Nonlinear Signal
  Alignment}.
\newblock In {\em Advances in Neural Information Processing Systems (NIPS)},
  2011.

\bibitem{CMU:mocap}
Carnegie Mellon~Graphics Lab.
\newblock {Motion Capture Database }.
\newblock \url{"http://mocap.cs.cmu.edu/info.php}, 2016.

\bibitem{Larsen:2006}
F.H. Larsen, F.~Van Den~Berg, and S.B. Engelsen.
\newblock An exploratory chemometric study of 1h nmr spectra of table wines.
\newblock {\em Journal of Chemometrics}, 20(5), 2006.

\bibitem{Lawrence:2005vk}
N.~D. Lawrence.
\newblock {Probabilistic Non-Linear Principal Component Analysis with Gaussian
  Process Latent Variable Models}.
\newblock {\em Journal of Machine Learning Research (JMLR)}, 6:1783--1816,
  2005.

\bibitem{Lawrence:2007hierarchical}
N~D Lawrence and A~J Moore.
\newblock Hierarchical gaussian process latent variable models.
\newblock In {\em Proceedings of the 24th international conference on Machine
  learning}, pages 481--488. ACM, 2007.

\bibitem{Lazaro:2012}
M.~L\'{a}zaro-Gredilla.
\newblock {Bayesian Warped Gaussian Processes}.
\newblock In {\em Advances in Neural Information Processing Systems (NIPS)}.
  2012.

\bibitem{Listgarten:2004}
J.~Listgarten, R.~M. Neal, S.~T. Roweis, and A.~Emili.
\newblock {Multiple Alignment of Continuous Time Series}.
\newblock In {\em Advances in Neural Information Processing Systems (NIPS)}.
  2005.

\bibitem{Lorbert:2012}
A.~Lorbert and P.~J. Ramadge.
\newblock {Kernel Hyperalignment}.
\newblock In {\em Advances in Neural Information Processing Systems (NIPS)}.
  2012.

\bibitem{Marron:2015}
J.S. Marron, J.O. Ramsay, L.M. Sangalli, and A.~Srivastava.
\newblock Functional data analysis of amplitude and phase variation.
\newblock {\em Statistical Science}, 30(4):468--484, 2015.

\bibitem{McCall:2012}
Corey McCall, Kishore~K. Reddy, and Mubarak Shah.
\newblock Macro-class selection for hierarchical k-nn classification of
  inertial sensor data.
\newblock In {\em PECCS}, 2012.

\bibitem{Muller:2007}
M.~M\"{u}ller.
\newblock {\em Information Retrieval for Music and Motion}.
\newblock 2007.

\bibitem{Rasmussen:2005}
C.~E. Rasmussen and C.~K.~I. Williams.
\newblock {\em {Gaussian Processes for Machine Learning (Adaptive Computation
  and Machine Learning)}}.
\newblock 2005.

\bibitem{Riihimaki:2010}
Jaakko Riihimäki and Aki Vehtari.
\newblock Gaussian processes with monotonicity information.
\newblock In Yee~Whye Teh and Mike Titterington, editors, {\em Proceedings of
  the Thirteenth International Conference on Artificial Intelligence and
  Statistics}, volume~9 of {\em Proceedings of Machine Learning Research},
  pages 645--652, Chia Laguna Resort, Sardinia, Italy, 13--15 May 2010. PMLR.

\bibitem{Snelson:2004}
E.~Snelson, Z.~Ghahramani, and C.~E. Rasmussen.
\newblock {Warped Gaussian Processes}.
\newblock In {\em Advances in Neural Information Processing Systems (NIPS)},
  2004.

\bibitem{Srivastava:2011}
A.~{Srivastava}, W.~{Wu}, S.~{Kurtek}, E.~{Klassen}, and J.~S. {Marron}.
\newblock {Registration of Functional Data Using Fisher-Rao Metric}.
\newblock {\em ArXiv}, abs/1103.3817, 2011.

\bibitem{SRVF:implement}
A.~{Srivastava}, W.~{Wu}, S.~{Kurtek}, E.~{Klassen}, and J.~S. {Marron}.
\newblock {Elastic Functional Data Analysis}.
\newblock \url{http://ssamg.stat.fsu.edu/software}, 2018.

\bibitem{Titsias:2009}
Michalis Titsias.
\newblock Variational learning of inducing variables in sparse gaussian
  processes.
\newblock In David van Dyk and Max Welling, editors, {\em Proceedings of the
  Twelth International Conference on Artificial Intelligence and Statistics},
  volume~5 of {\em Proceedings of Machine Learning Research}, pages 567--574,
  Hilton Clearwater Beach Resort, Clearwater Beach, Florida USA, 16--18 Apr
  2009. PMLR.

\bibitem{Trigeorgis:2016}
G.~Trigeorgis, M.~A. Nicolaou, S.~Zafeiriou, and B.~W. Schuller.
\newblock {Deep Canonical Time Warping}.
\newblock In {\em {International Conference on Computer Vision and Pattern
  Recognition (CVPR)}}, 2016.

\bibitem{Trigeorgis:2017}
G.~Trigeorgis, M.~A. Nicolaou, S.~Zafeiriou, and B.~W. Schuller.
\newblock {Deep Canonical Time Warping for Simultaneous Alignment and
  Representation Learning of Sequences}.
\newblock {\em IEEE Transactions on Pattern Analysis and Machine Intelligence
  (PAMI)}, 2017.

\bibitem{Tucker:2013}
J.~D. Tucker, W.~Wu, and A.~Srivastava.
\newblock {Generative Models for Functional Data using Phase and Amplitude
  Separation}.
\newblock {\em Computational Statistics and Data Analysis}, 61(Supplement
  C):50--66, 2013.

\bibitem{Urtasun:2005}
R.~Urtasun, D.~J. Fleet, A.~Hertzmann, and P.~Fua.
\newblock {Priors for People Tracking from Small Training Sets}.
\newblock In {\em International Conference on Computer Vision (ICCV)}, 2005.

\bibitem{Vu:2012}
H.~T. Vu, C.~J. Carey, and S.~Mahadevan.
\newblock {Manifold Warping: Manifold Alignment over Time}.
\newblock 2012.

\bibitem{Yu:2009}
Byron~M Yu, John~P Cunningham, Gopal Santhanam, Stephen~I. Ryu, Krishna~V
  Shenoy, and Maneesh Sahani.
\newblock Gaussian-process factor analysis for low-dimensional single-trial
  analysis of neural population activity.
\newblock In D.~Koller, D.~Schuurmans, Y.~Bengio, and L.~Bottou, editors, {\em
  Advances in Neural Information Processing Systems 21}, pages 1881--1888.
  Curran Associates, Inc., 2009.

\bibitem{Zhou:2012}
F.~Zhou.
\newblock {Generalized Time Warping for Multi-modal Alignment of Human Motion}.
\newblock In {\em International Conference on Computer Vision and Pattern
  Recognition (CVPR)}, 2012.

\bibitem{Zhou:2016}
F.~Zhou and F.~{de al Torre}.
\newblock {Generalized Canonical Time Warping}.
\newblock {\em IEEE Transactions on Pattern Analysis and Machine Intelligence
  (PAMI)}, 38(2), 2016.

\bibitem{Zhou:2009}
F.~Zhou and F.~{de la Torre}.
\newblock {Canonical Time Warping for Alignment of Human Behavior}.
\newblock In {\em Advances in Neural Information Processing Systems (NIPS)},
  2009.

\bibitem{CTW:implement}
F.~Zhou and F.~{de la Torre}.
\newblock {Software for Canonical Time Warping}.
\newblock \url{http://www.f-zhou.com/ta_code.html}, 2018.

\end{thebibliography}
